%% file: _main.tex
\input{_constants}
\arxiv 

\pdfoutput=1
\documentclass[10pt,twocolumn,letterpaper]{article}
\usepackage{svg}
\input{cvpr_header}

\unless\ifarxiv \myexternaldocument{_supplementary} \fi

\begin{document}
\title{\paperTitle}
\author{\authorBlock}
\maketitle

\input{00_abstract}
\input{01_intro}

\input{02_related}

\input{03_method}

\input{04_experiments}

\input{10_conclusion}

{\small
\bibliographystyle{ieeenat_fullname}
\bibliography{11_references}
}

\ifarxiv \clearpage \appendix \input{12_appendix} \fi

\end{document}

%% file: _constants.tex
\def\paperID{16005}
\def\confName{CVPR}
\def\confYear{2026}

\def\paperTitle{Deconstructing Generative Diversity: An Information Bottleneck Analysis of Discrete Latent Generative Models}

\def\authorBlock{
    Yudi Wu \\
    {\tt\small Zhejiang University} \\
    {\tt\small wuyudi@zju.edu.cn}
    \and
    Wenhao Zhao \\
    {\tt\small National University of Singapore}\\
    {\tt\small wenhaozhao@u.nus.edu}
    \and
    Dianbo Liu \\
    {\tt\small National University of Singapore} \\
    {\tt\small dianbo@nus.edu.sg}
}

\newif\ifreview 
\newif\ifarxiv \newcommand{\arxiv}{\arxivtrue}
\newif\ifcamera 
\newif\ifrebuttal 

%% file: cvpr_header.tex
\ifreview \usepackage[review]{cvpr} \fi
\ifarxiv \usepackage[pagenumbers]{cvpr} \fi
\ifrebuttal \usepackage[rebuttal]{cvpr} \fi
\ifcamera \usepackage{cvpr} \fi

\input{_macros}  

\usepackage{xr-hyper}

\makeatletter
\newcommand*{\addFileDependency}[1]{
  \typeout{(#1)}
  \@addtofilelist{#1}
  \IfFileExists{#1}{}{\typeout{No file #1.}}
}

\makeatother
\newcommand*{\myexternaldocument}[1]{
    \externaldocument{#1}
    \addFileDependency{#1.tex}
    \addFileDependency{#1.aux}
}

\definecolor{cvprblue}{rgb}{0.21,0.49,0.74}
\usepackage[pagebackref,breaklinks,colorlinks,allcolors=cvprblue]{hyperref}
\usepackage[capitalize]{cleveref}
\crefname{section}{Sec.}{Secs.}
\crefname{table}{Table}{Tables}
\crefname{figure}{Fig.}{Figs.}

\ifarxiv \crefname{appendix}{App.}{Apps.}
\else \crefname{appendix}{Suppl.}{Suppls.} \fi

%% file: _macros.tex
\usepackage{graphicx}	
\usepackage{amsmath}	
\usepackage{amssymb}	
\usepackage{booktabs}
\usepackage{times}
\usepackage{microtype}
\usepackage{epsfig}
\usepackage{caption}
\usepackage{float}
\usepackage{placeins}
\usepackage{color, colortbl}
\usepackage{stfloats}
\usepackage{enumitem}
\usepackage{tabularx}
\usepackage{xstring}
\usepackage{multirow}
\usepackage{xspace}
\usepackage{url}
\usepackage{subcaption}
\usepackage{xcolor}
\usepackage[hang,flushmargin]{footmisc}

\ifcamera \usepackage[accsupp]{axessibility} \fi

% ...

\ifarxiv  \fi

\newcommand{\R}[1]{{%
    \textbf{%
        \ifstrequal{#1}{1}{\textcolor{red}{R#1}}{%
        \ifstrequal{#1}{2}{\textcolor{blue}{R#1}}{%
        \ifstrequal{#1}{3}{\textcolor{magenta}{R#1}}{%
        \ifstrequal{#1}{4}{\textcolor{teal}{R#1}}{%
                           \textcolor{cyan}{R#1}%
        }}}}%
    }%
}}

%% file: 00_abstract.tex
\begin{abstract}
Generative diversity varies significantly across discrete latent generative models such as AR, MIM, and Diffusion. We propose a diagnostic framework, grounded in Information Bottleneck (IB) theory, to analyze the underlying strategies resolving this behavior. The framework models generation as a conflict between a 'Compression Pressure'—a drive to minimize overall codebook entropy—and a 'Diversity Pressure'—a drive to maximize conditional entropy given an input. We further decompose this diversity into two primary sources: 'Path Diversity', representing the choice of high-level generative strategies, and 'Execution Diversity', the randomness in executing a chosen strategy. To make this decomposition operational, we introduce three zero-shot, inference-time interventions that directly perturb the latent generative process and reveal how models allocate and express diversity. Application of this probe-based framework to representative AR, MIM, and Diffusion systems reveals three distinct strategies: “Diversity-Prioritized” (MIM), “Compression-Prioritized” (AR), and “Decoupled” (Diffusion). Our analysis provides a principled explanation for their behavioral differences and informs a novel inference-time diversity enhancement technique.

\end{abstract}

%% file: 01_intro.tex
\section{Introduction}
\label{sec:intro}

Discrete latent generative models have recently emerged as a central paradigm in image synthesis. Approaches based on vector quantization (VQ)\cite{vqvae}, such as autoregressive transformers\cite{llamagen}, masked image models\cite{patil2024amusedopenmusereproduction}, and diffusion models operating in discrete latent spaces\cite{vqdiff}, have demonstrated remarkable progress in controllable and high-fidelity generation. The discrete formulation offers compact and structured representations that align well with token-based learning and scalable training pipelines. As these models become increasingly prevalent, understanding their underlying generative behavior—particularly the nature and source of their \textit{diversity}—has become a critical research question.

Generative models have been widely studied from the perspective of evaluating or quantifying their output variation. Recent works have proposed a range of metrics to assess novelty, variability, or originality in generated samples~\cite{groupcreativity,infoDiver,diversityIssue}. These studies have improved our ability to measure how diverse model outputs appear but provide limited insight into the mechanisms that produce such variation. In particular, they seldom address how different model architectures internalize and control stochasticity within their latent representations. As a result, we still lack a systematic understanding of why discrete latent models such as autoregressive, masked, and diffusion-based frameworks exhibit distinct patterns of generative behavior.

We address this gap through an information-theoretic framework grounded in the Information Bottleneck (IB) principle. From this perspective, generative diversity arises from the balance between two opposing pressures: a \textit{compression pressure}, which encourages compact latent representations and low entropy, and a \textit{diversity pressure}, which promotes stochastic and expressive mappings that retain uncertainty. We further decompose this diversity into interpretable components that correspond to variability in generative strategy (\textit{path diversity}) and randomness during execution (\textit{execution diversity}), providing a unified view of information flow in generation.

To operationalize this analysis, we develop a set of zero-shot, inference-time probes that directly perturb a model’s latent generative process. Each probe targets a different component of the IB decomposition—codebook usage, sampling stochasticity, and prompt conditioning—and measures how the model’s outputs change under these controlled interventions. By examining a model’s sensitivity to these perturbations, we expose how its internal mechanism navigates the trade-off between compression and diversity. 

The resulting probe responses reveal several recurring patterns across architectures. Some models behave as \emph{compression-prioritized} systems, showing minimal changes under perturbations and consistently producing stable, low-variance outputs. Others are \emph{diversity-prioritized}, maintaining high conditional entropy and expressing substantial variation even when constraints are imposed. A third group exhibits \emph{decoupled} behavior, where path-level randomness and execution-level randomness contribute independently, yielding models that remain stable at the structural level while preserving controlled variation during sampling. These patterns provide a coherent view of how different discrete generative models manage information flow and where their diversity ultimately originates.

Our contributions are summarized as follows:
\begin{enumerate}
    \item We propose a unified information-theoretic framework for analyzing generative diversity in discrete latent models. 
    \item We conduct a systematic study of representative model architectures to reveal how they resolve the trade-off between compression and diversity. 
    \item We demonstrate a simple inference-time method for enhancing diversity in pretrained models without retraining, offering a practical tool for controlling generative behavior.
\end{enumerate}

%% file: 02_related.tex
\section{Related Work}
\label{sec:related}
\input{figs/Figure1}

\subsection{Discrete Latent Generative Models}
Discrete latent variable models, such as VQ-VAE\cite{vqvae}, form the foundation of modern image generation by introducing a codebook that discretizes continuous representations into symbolic latent tokens. Subsequent developments, including VQ-GAN\cite{VQGAN} and VQ-VAE 2\cite{vqave2}, have shown that discretization stabilizes training and improves sample fidelity. Despite sharing the same underlying discrete latent space formulation, different model families employ distinct \emph{generation strategies} within this space. Autoregressive (AR) models\cite{pixelcnn,var,llamagen} produce tokens sequentially in a fixed left-to-right order, Masked Image Models (MIMs)\cite{maskgit,muse,show-o} iteratively predict and reveal masked regions following a stochastic unmasking schedule, while diffusion-based models\cite{LDM,vqdiff} reconstruct the latent code through a noise-to-data denoising trajectory. In other words, their differences lie not in the representational domain itself, but in \emph{how they traverse and utilize the discrete latent space during generation}. These distinct traversal mechanisms naturally lead to different diversity behaviors: AR models tend to be compression-prioritized and produce deterministic outputs; MIMs introduce strong stochasticity through masking and exhibit diversity-prioritized behavior; and diffusion models decouple semantic consistency from sampling stochasticity, achieving path-driven diversity. Prior works have largely treated these differences as architectural, without a principled theoretical explanation. Our work instead provides an information-theoretic framework that unifies these generation paradigms by interpreting their behavioral differences as distinct resolutions of the Information Bottleneck trade-off between compression and diversity.

\subsection{Information Bottleneck in Representation and Generation}
The Information Bottleneck (IB) principle~\cite{ib} provides a theoretical framework for understanding how learning systems balance compression and informativeness. Variational extensions~\cite{vib} and adaptations for discrete latent models~\cite{vibvq} show that the mutual information among inputs, latent variables, and outputs determines how representations retain relevant content while discarding redundancy. In generative modeling, this trade-off can be viewed as a balance between reducing latent complexity and preserving sufficient information to produce diverse and faithful samples. The IB perspective thus offers a unified view of how generative models regulate information flow between input prompts, latent codes, and outputs.

\subsection{Quantization Schemes as Information Constraints}
Quantization plays a central role in defining the information flow of discrete latent models. Traditional VQ approaches rely on learned codebooks that often suffer from codebook collapse\cite{vqvae}. Recent advances propose alternative quantization schemes that impose structural constraints rather than learned ones. Finite Scalar Quantization (FSQ)\cite{FSQ} employs a fixed Cartesian grid, ensuring full utilization of the codebook by design; Lookup-Free Quantization (LFQ)\cite{LFQ} simplifies this to binary levels, embodying a structural compression strategy; and Binary Spherical Quantization (BSQ)~\cite{BSQ} further introduces geometric normalization to bound quantization error. While these schemes were proposed independently, they can be interpreted as distinct instantiations of the IB optimization under different quantization constraints. Our framework unifies these perspectives, showing that each scheme implicitly prioritizes a specific resolution of the compression–diversity trade-off.

%% file: figs/Figure1.tex
\begin{figure*}[tp]
    \centering
    \includegraphics[width=\textwidth]{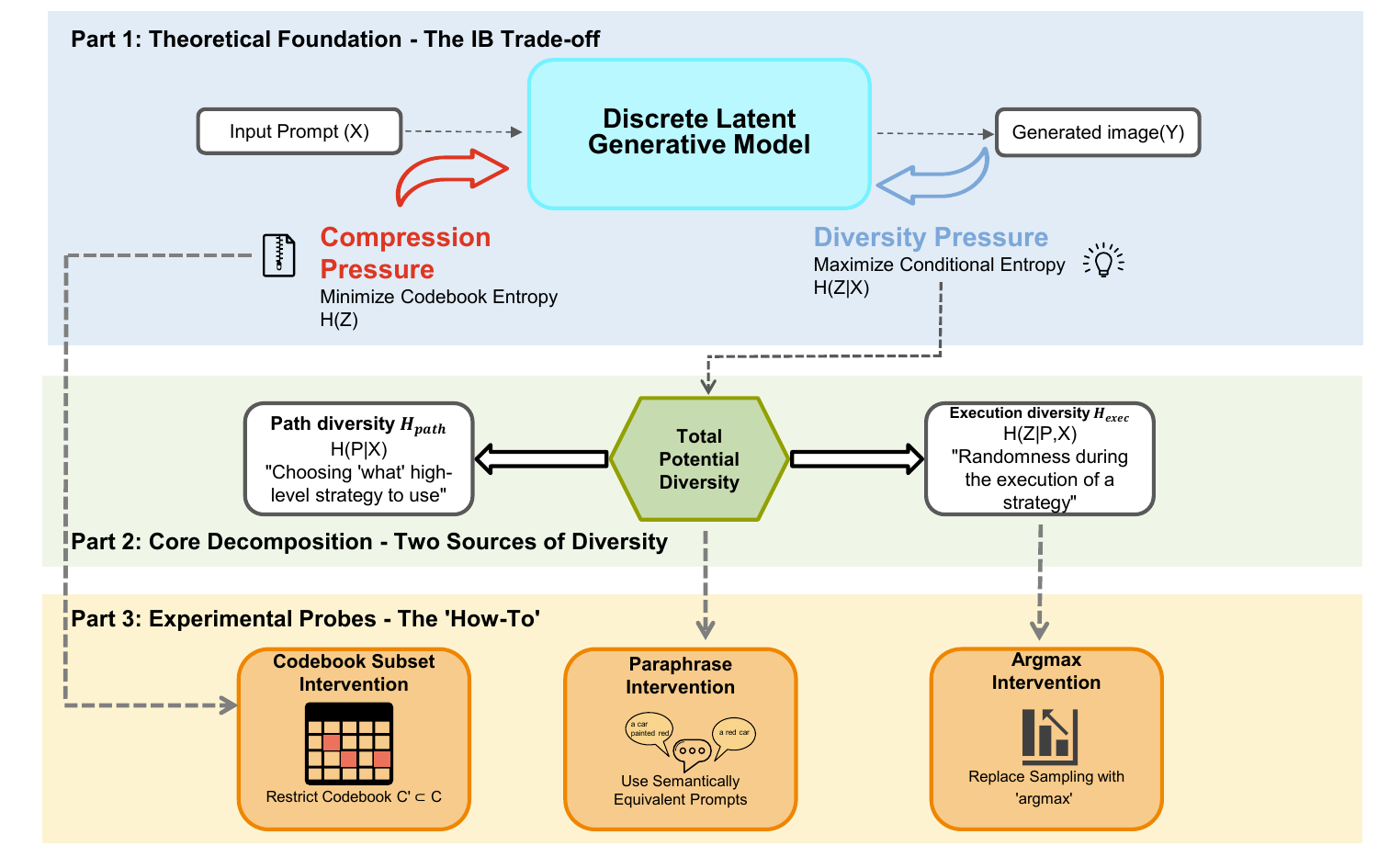}
    \caption{Conceptual overview of our diagnostic framework. \textbf{Part 1: Theoretical Foundation.} Our framework is grounded in the Information Bottleneck (IB) principle, which imposes two conflicting pressures on any VQ-based model: a \textbf{Compression Pressure} to minimize codebook entropy $H(Z)$ and a \textbf{Diversity Pressure} to maximize conditional entropy $H(Z|X)$. \textbf{Part 2: Core Decomposition.} We refine the concept of diversity by decomposing $H(Z|X)$ into two distinct sources: \textbf{Path Diversity ($H_{path}$)}, which represents the choice of high-level generative strategies, and \textbf{Execution Diversity ($H_{exec}$)}, which represents the lower-level randomness in executing a chosen strategy. \textbf{Part 3: Experimental Probes.} We introduce three zero-shot interventions to diagnose how a model resolves the IB conflict. The `Codebook Subset` intervention probes the model's response to the Compression Pressure and measures $H(Z)$. The `Argmax` and `Paraphrase` interventions serve as complementary probes to measure Execution Diversity $H(Z|P,X)$ and the overall magnitude of $H(Z|X)$, respectively.} 
    \label{fig:1}
\end{figure*}

%% file: 03_method.tex
\section{Method}
\label{sec:method}

We present a diagnostic framework, grounded in the Information Bottleneck (IB) perspective, to analyze generative diversity in VQ-based models. The framework aims to disentangle and quantify distinct sources of diversity in pre-trained models via zero-shot, inference-time interventions. We summarize the theoretical view in \cref{theory} and detail the probes in \cref{exp_probes}.
\subsection{ Defining Generative Diversity}
In this study, we regard generative diversity as the ability of a model to produce a wide range of distinct yet high-quality outputs under the same semantic conditions. A model with high diversity should be capable of exploring multiple valid realizations of a concept while maintaining fidelity and coherence in each instance. 

\subsection{Theoretical framework}
\label{theory}

Let \(X\) be the input prompt, \(Z\) the discrete latent code produced by a VQ encoder, and \(Y\) the generated image.  
From the Information Bottleneck (IB) perspective~\cite{ib,vibvq}, a generative model learns a latent representation \(Z\) that balances two competing objectives: it should compress the input \(X\) as much as possible while still preserving the information relevant for generating \(Y\).  
This trade-off can be formalized by the IB Lagrangian:
\begin{equation}
\mathcal{L}_{\text{IB}} = I(X;Z) - \beta I(Z;Y)
\label{eq:ib_objective}
\end{equation}
where \(I(\cdot;\cdot)\) denotes mutual information and \(\beta\) controls the relative strength of the two terms.  
The first term encourages compression by minimizing the dependency between \(X\) and \(Z\), while the second term promotes expressiveness by maximizing the information that \(Z\) retains about \(Y\).  

For our diagnostic purposes, it is sufficient to work with Shannon entropy \(H(\cdot)\) and conditional entropy \(H(\cdot|\cdot)\), since mutual information decomposes as
\begin{equation}
I(X;Z)=H(Z)-H(Z|X)
\label{eq:ib_basic}
\end{equation}
This expression itself exposes the core tension that shapes the behavior of generative models. The term \(H(Z)\) measures the overall uncertainty of the latent space, and minimizing it encourages compact, low-entropy representations. In contrast, \(H(Z|X)\) quantifies the stochasticity of the latent code given an input and grows when multiple valid latent outcomes exist for the same prompt. Reducing \(I(X;Z)\) thus requires decreasing \(H(Z)\) while increasing \(H(Z|X)\), enforcing a balance between representation compactness and conditional variability. The way a model resolves this internal conflict determines its characteristic pattern of generative diversity.

To analyze how this variability manifests during the generation process, we introduce \(P\) to denote the generative \emph{path}—the high-level sequence of operations that produces \(Z\). The exact form of \(P\) depends on the model’s generation strategy, and the total conditional uncertainty of \(Z\) given \(X\) can be further decomposed via the law of total entropy:

\begin{equation}
H(Z|X)=H(P|X)+H(Z|P,X)-H(P|Z,X)
\label{eq:total_entropy}
\end{equation}
The two forward-looking terms admit a meaningful upper bound on \(H(Z|X)\); we therefore focus on their contributions:
\begin{itemize}
    \item \textbf{Path diversity:} \(H_{\text{path}} \triangleq H(P|X)\), the entropy of the chosen high-level generative strategies.
    \item \textbf{Execution diversity:} \(H_{\text{exec}} \triangleq H(Z|P,X)\), the entropy remaining when a path is fixed.
\end{itemize}
These quantities provide a compact decomposition of potential generative variability without introducing additional notation. In practice, the residual term \(H(P|Z,X)\) is typically small when \(P\) is recoverable from \(Z\); we discuss its role when relevant.

Below we characterize the three generation strategies considered in this work in terms of \(P\):
\begin{itemize}
    \item \textbf{Autoregressive (AR):} \(P\) is the deterministic sequential order of generation,
    \begin{equation}
    P = (1,2,\dots,N), \quad p(P|X)=1
    \label{eq:ar_path}
    \end{equation}
    implying \(H_{\text{path}}=0\). All diversity arises from stochastic token sampling conditioned on \(P\), reflected in \(H_{\text{exec}}>0\).

    \item \textbf{Masked image models (MIM):} \(P\) is the ordered sequence of masking states over \(T\) steps,
    \begin{equation}
    P = (M_1, M_2, \dots, M_T), \quad M_t \subseteq \{1,\dots,N\}
    \label{eq:mim_path}
    \end{equation}
    where \(M_t\) denotes the set of masked token indices at step \(t\), with \(M_1=\{1,\dots,N\}\) and \(M_{t+1}\) obtained by removing a stochastic subset of indices from \(M_t\) according to a deterministic masking schedule \(\gamma(t/T)\).  
    The distribution \(p(P|X)\) is therefore induced by random unmasking and has positive entropy:
    \begin{equation}
    H_{\text{path}} = H(P|X) = -\mathbb{E}_{p(P|X)}[\log p(P|X)] > 0
    \label{eq:mim_entropy}
    \end{equation}
    Execution diversity arises from token sampling within each step, contributing to \(H_{\text{exec}}\).

    \item \textbf{Diffusion models:} \(P\) is the latent trajectory of length \(T\),
    \begin{equation}
\begin{split}
P &= (z_T, z_{T-1}, \dots, z_1), \\
p(P|X) &= p(z_T)\prod_{t=1}^{T-1} p_\theta(z_t|z_{t+1},X)
\end{split}
\label{eq:diff_path}
\end{equation}
    where \(z_T \sim \mathcal{N}(0,I)\) is the initial noise and each transition \(p_\theta(z_t|z_{t+1},X)\) is typically Gaussian with learned mean and variance.  
    The path entropy is thus
    \begin{equation}
    H_{\text{path}} = H(P|X) = H(z_T) + \sum_{t=1}^{T-1} H(z_t|z_{t+1},X)
    \label{eq:diff_entropy}
    \end{equation}
    which is large due to randomness accumulated from initialization and stochastic denoising. The conditional entropy \(H_{\text{exec}}\) is typically smaller since generation becomes nearly deterministic once the initial noise is fixed.
\end{itemize}

\subsection{Quantization schemes as IB constraints}
\label{sec:general_quant}

The quantization scheme that defines \(Z\) constrains the IB trade-off by shaping the latent domain and its entropy. We summarize three representative schemes using consistent notation.

\textbf{Finite Scalar Quantization (FSQ).\cite{FSQ}} Project encoder output to \(\mathbb{R}^d\) and quantize each dimension into \(L_i\) levels. The implicit codebook \(\mathcal{C}\) has size$|\mathcal{C}|=\prod_i L_i$,
and \(H(Z)\) measures utilization of this fixed grid.

\textbf{Lookup-Free / Binary FSQ (LFQ).\cite{LFQ}} A binary instance of FSQ with \(L_i=2\). The codebook is structurally bounded and \(H(Z)\) limited by design; diversity must be realized via conditional entropy \(H(Z|X)\).

\textbf{Binary Spherical Quantization (BSQ).\cite{BSQ}} Apply \(\ell_2\)-normalization prior to binary quantization, effectively partitioning a unit hypersphere. The geometric constraint alters quantization error and the form of \(H(Z)\).

Treating these schemes uniformly under the IB view clarifies how quantization geometry biases models toward different balances between compression and diversity.

\subsection{Experimental probes}
\label{exp_probes}

We design three zero-shot, inference-time interventions that require no retraining. Each probe isolates a distinct term in our decomposition and provides an interpretable signal of the model’s generative behavior.

\subsubsection{Codebook subset}
\label{probe:subset}
This probe examines how strongly a model depends on its codebook entropy \(H(Z)\). We limit the available entries to a smaller subset \(\mathcal{C}' \subset \mathcal{C}\), chosen according to usage frequency on a validation set, and generate outputs under this reduced vocabulary. By comparing the diversity of results before and after restriction, we can infer the model’s reliance on codebook capacity. A notable reduction in diversity indicates sensitivity to compression, whereas similar diversity levels suggest effective operation within a compact latent space.

\subsubsection{Argmax (deterministic sampling)}
\label{probe:argmax}
To assess execution-level randomness \(H_{\text{exec}}\), stochastic sampling during inference is replaced with deterministic argmax decoding, while other random factors such as initialization or masking remain unchanged. Observing the change in diversity after this substitution reveals the role of sampling noise in generation. When diversity declines sharply, the model relies heavily on token-level randomness; when diversity remains steady, most variation originates from higher-level path choices.

\subsubsection{Paraphrase set}
\label{probe:paraphrase}
Finally, we probe the conditional entropy \(H(Z|X)\) by evaluating the model’s response to semantic variations in the input. Several paraphrased prompts with equivalent meaning but different phrasing are used to produce outputs, which are then aggregated for analysis. If the resulting images differ widely, the model distributes semantically similar inputs across a broad region of the latent space, indicating flexible and expressive mapping. If the outputs remain consistent, the model encodes semantics in a narrow, low-entropy region, reflecting conservative generation behavior.

Each probe therefore provides complementary evidence about how a model resolves the trade-off between compression and diversity, distinguishing whether its diversity is primarily path-driven, execution-driven, or limited by codebook capacity.

%% file: 04_experiments.tex
\section{Experiments}
\label{sec:experiment}
\input{figs/disentangle}
\input{figs/Figure5}
\input{figs/Figure6}
\input{figs/ablations}

This section compares three distinct model architectures—Autoregressive (AR), Masked Image Modeling (MIM), and Diffusion—to demonstrate the effectiveness of our diagnostic framework. We show that their well-known but poorly understood behavioral differences can be explained as three unique, archetypal strategies for resolving the IB conflict. Further experiments could be found in our Appendix.
\subsection{Experimental Setup}
\begin{itemize}
    \item \textbf{Models}: We select representative open-source models for each of the three VQ-based sequential architectures: \textbf{LlamaGen (AR)}\cite{llamagen}, \textbf{aMUSEd (MIM)}\cite{patil2024amusedopenmusereproduction}, and \textbf{VQ-Diffusion(Diffusion)}\cite{vqdiff}. All models are evaluated on their publicly available pre-trained checkpoints without any fine-tuning.
    \item \textbf{Dataset \& Metrics}: We use a high-quality synthetic dataset of captions generated by DALL-E 3\cite{Egan_Dalle3_1_Million_2024} together with MSCOCO2014 dataset to calculate FID. Generative diversity is quantified using LPIPS\cite{LPIPS}, pixel cosine distance, and SSIM\cite{ssim}. Following established protocols, for each input prompt, we generate a set of output images. The diversity for that prompt is calculated by averaging the metric score over all unique pairs of the generated images. The final reported diversity value is the mean of these per-prompt scores across the entire test set.
    \item \textbf{Interventions}: For each model, we apply the three experimental probes described in \cref{exp_probes} The paraphrase test utilizes a set of 5 paraphrased versions for each of base prompts, generated by GPT-5.
\end{itemize}

\subsection{Comparative Analysis of Model Strategies}
\label{main_exp}
Our framework reveals three distinct strategies for resolving the IB conflict, perfectly aligning with the experimental evidence from the three model classes.
\subsubsection{MIM (aMUSEd): The Diversity-Prioritized Strategy}
As shown in \cref{fig:amused_disentangle}, the aMUSEd model exhibits a significant drop in diversity after both the Argmax and Subset interventions. This demonstrates a strategy that heavily prioritizes diversity. The significant drop from the Argmax intervention shows high Execution Diversity ($H_{exec}$), and the Paraphrase Intervention confirms this: the significant increase in diversity provides cross-validating evidence of a high overall $H(Z|X)$. To support this high level of stochasticity, the model must resist compression, which is confirmed by its significant Subset drop (high $H(Z)$).
\subsubsection{AR (LlamaGen): The Compress-Prioritized Strategy}
The LlamaGen model, as shown in \cref{fig:llamagen_disentangle}, presents a unique generative profile. The Argmax intervention reduces all diversity metrics to zero (not plotted), confirming that its generative diversity stems entirely from stochastic token sampling ($H_{exec}$), meaning its Path Diversity ($H_{path}$) is effectively zero. Furthermore, the Subset codebook intervention results in only a minor drop in diversity. According to our framework, this pair of findings points to a strategy that heavily prioritizes compression, leading the model to rely on a small, low-entropy subset of its codebook—a state of effective codebook collapse. Intriguingly, and in stark contrast to this compressed state, the model is highly sensitive to input variations. The Paraphrase intervention leads to an increase across all diversity metrics. This reveals that while the model has a collapsed vocabulary (low $H(Z)$), its mapping from prompt to latent space is not semantically robust; it maps syntactic variations of the same concept to distinct regions within its limited latent space.  
\subsubsection{VQ-Diffusion(Diffusion): The Decoupled Strategy}
The VQ-Diffusion model presents a more sophisticated case, as shown in \cref{fig:vqdiff_disentangle}. It shows a non-significant difference in diversity after the Argmax intervention, yet a significant drop after the Subset intervention. Our framework resolves this apparent contradiction. The results indicate a strategy that relies on high Path Diversity ($H_{path}$) rather than Execution Diversity ($H_{exec}$) for its final diversity. Its stability under paraphrase is not a sign of collapse, but of robustness; its structured generative path effectively filters syntactic noise, making it a near-perfect IB that successfully decouples semantic understanding from creative execution.

\subsection{Impact of Interventions on Generation Quality}
Beyond diversity, we also examine how the three experimental probes influence generation quality. As shown in \cref{fig:5}, the effects differ across models. For aMUSEd, both the Argmax and Subset interventions lead to visible declines in CLIP and IQA scores, mirroring their strong reduction in diversity and indicating that its generation quality depends heavily on stochastic sampling and rich codebook usage. LlamaGen, by contrast, maintains stable performance under all interventions, with minimal variation in CLIP, IQA, or FID, reflecting a deterministic and tightly constrained generation process. VQ-Diffusion shows little sensitivity to the Argmax or Paraphrase probes, consistent with its path-driven diversity, but experiences a clear drop in IQA when codebook access is restricted. This suggests that while its creative variability is primarily structural, high-quality synthesis still requires the full representational capacity of the latent vocabulary.
\input{figs/VLM}
\input{figs/samples}
\subsection{Ablation Studies}
\label{sec:ablations}
We conduct ablations to examine model sensitivity to codebook capacity and to validate probe settings.
\subsubsection{Impact of Codebook Subset Ratio}
\label{sec:ablation_subset_ratio}
To analyze dependence on codebook capacity, we vary the subset ratio and record LPIPS diversity and CLIP-IQA scores (\cref{fig:6}).  
The diversity trends (\cref{fig:6L}) show that aMUSEd’s diversity rises sharply with larger codebooks, LlamaGen saturates quickly, and VQ-Diffusion exhibits gradual, resilient behavior.  
The quality trends (\cref{fig:6R}) show consistent degradation under severe compression for all models.  
The subset ratio used in the main experiments (\cref{main_exp}) corresponds to the inflection point where diversity is most sensitive but quality remains stable, ensuring that the Subset probe operates in the regime most informative for cross-model comparison.

\subsubsection{Temporal Sensitivity to Execution Stochasticity}
\label{sec:ablation_argmax_timing}

To understand \textit{when} execution diversity ($H_{exec}$) is most critical during aMUSEd's iterative generation process, we selectively applied the Argmax intervention to early, middle, or late stages. As shown in \cref{fig:argmax_timing}, the results indicate that stochasticity during the initial phases of generation is substantially more critical for establishing the final output diversity than randomness applied in later refinement stages. Applying Argmax early nearly replicates the diversity reduction seen under the fully deterministic condition.
\subsubsection{Influence of Prompt Length on Paraphrase Sensitivity}
\label{sec:ablation_prompt_length}

To examine whether generative model's sensitivity to input variations ($H(Z|X)$) is modulated by prompt complexity, proxied by length, we evaluated the effect of paraphrases of varying lengths. We generated outputs using sets of paraphrases categorized as short, medium, long, or a mix of all lengths. As detailed in \cref{fig:prompt_length}, while all paraphrase conditions increased diversity compared to the baseline, utilizing a mixture of prompts with varying lengths yielded the most substantial diversity enhancement. This suggests the model benefits most from exposure to diverse syntactic structures representing the same semantic concept.
\subsubsection{Codebook Reliance Structure via Alternative Subsetting}
\label{sec:ablation_subset_method}

The structure of codebook reliance ($H(Z)$) was further probed by comparing alternative methods for subset construction, beyond simply removing the least frequent vectors. Keeping the subset size fraction constant, we evaluated the impact of removing the \textit{most} frequent vectors and removing a \textit{random} selection of vectors. The results, shown in \cref{fig:subset_method}, reveal a surprising effect. While removing the least frequent vectors significantly reduces diversity, removing the \textit{most} frequent vectors leads to a substantial \textit{increase} in diversity compared to the baseline. Removing random vectors results in a moderate increase. This counter-intuitive finding suggests that forcing the model away from its preferred high-frequency codes compels it to explore less common but potentially more diverse combinations within its vocabulary, enhancing overall output variation without compromising quality.
\subsection{Proposed Inference Strategy for Diversity Enhancement}
\label{sec:proposed_method}

Motivated by the findings in our ablation studies (\cref{sec:ablation_prompt_length} and \cref{sec:ablation_subset_method}), particularly the observed diversity benefits from mixed-length paraphrases and the counter-intuitive diversity increase upon disabling high-frequency codebook tokens, we propose a novel inference-time strategy aimed at enhancing generative diversity without requiring model retraining.

This method combines two techniques applied during generation: 
\begin{enumerate}
    \item  Utilizing a set of mixed-length paraphrases as input prompts to leverage diverse syntactic structures representing the same concept.
    \item Disabling a pre-defined fraction of the most frequently used codebook tokens (identified via validation set statistics), forcing the model to explore less common representational pathways.
\end{enumerate}

To provide preliminary validation for this approach, we applied this combined intervention strategy during inference with pre-trained Vision-Language Models (VLM), DeepSeek Janus Pro 1B\cite{janus,januspro} and Show-o\cite{show-o,show-o2}, adapted for token-based visual generation. \cref{fig:vlm_diversity_enhancement} shows the LPIPS diversity and CLIP-IQA score as a function of the codebook subset ratio when applying this strategy, compared to the baseline generation. The results demonstrate a clear increase in LPIPS diversity (\cref{fig:Janus_lpips_curve} \cref{fig:show-o_lpips_curve}) compared to the baseline, while the CLIP-IQA score (\cref{fig:Janus_iqa_curve} \cref{fig:show-o_iqa_curve}) remains competitive. We provide some samples at \cref{fig:qualitative_grid} for reference. This suggests potential for developing zero-shot, inference-time techniques to modulate the diversity-fidelity trade-off in pre-trained generative models, guided by diagnostic insights into their underlying mechanisms.

%% file: figs/disentangle.tex
\begin{figure*}[tp]
    \centering

    \begin{subfigure}[b]{0.33\textwidth}
        \centering
        \includegraphics[width=\linewidth]{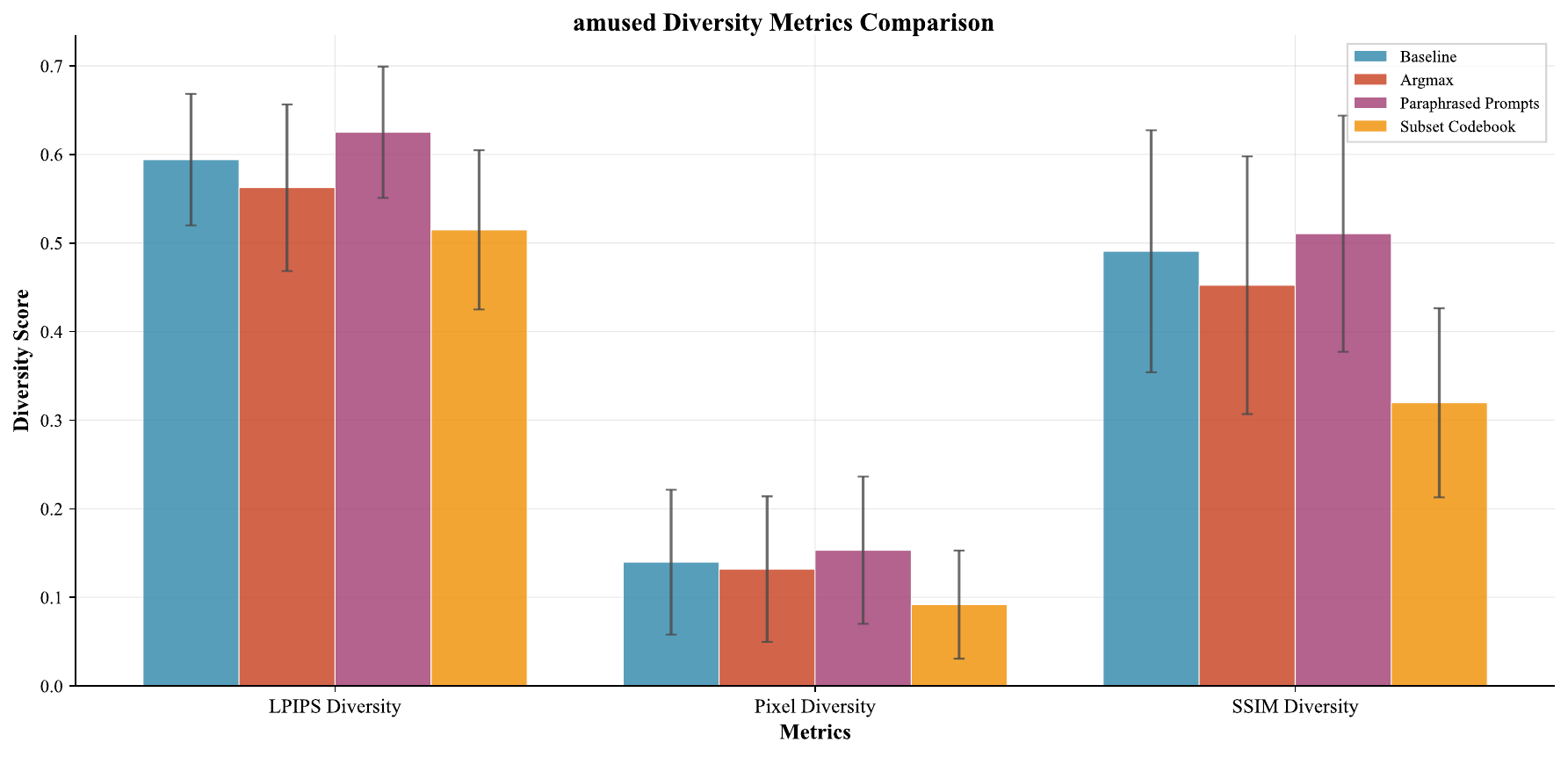}
        \caption{aMUSEd Diversity Comparsion}
        \label{fig:amused_disentangle}
    \end{subfigure}
    \hfill 
    \begin{subfigure}[b]{0.33\textwidth}
        \centering
        \includegraphics[width=\linewidth]{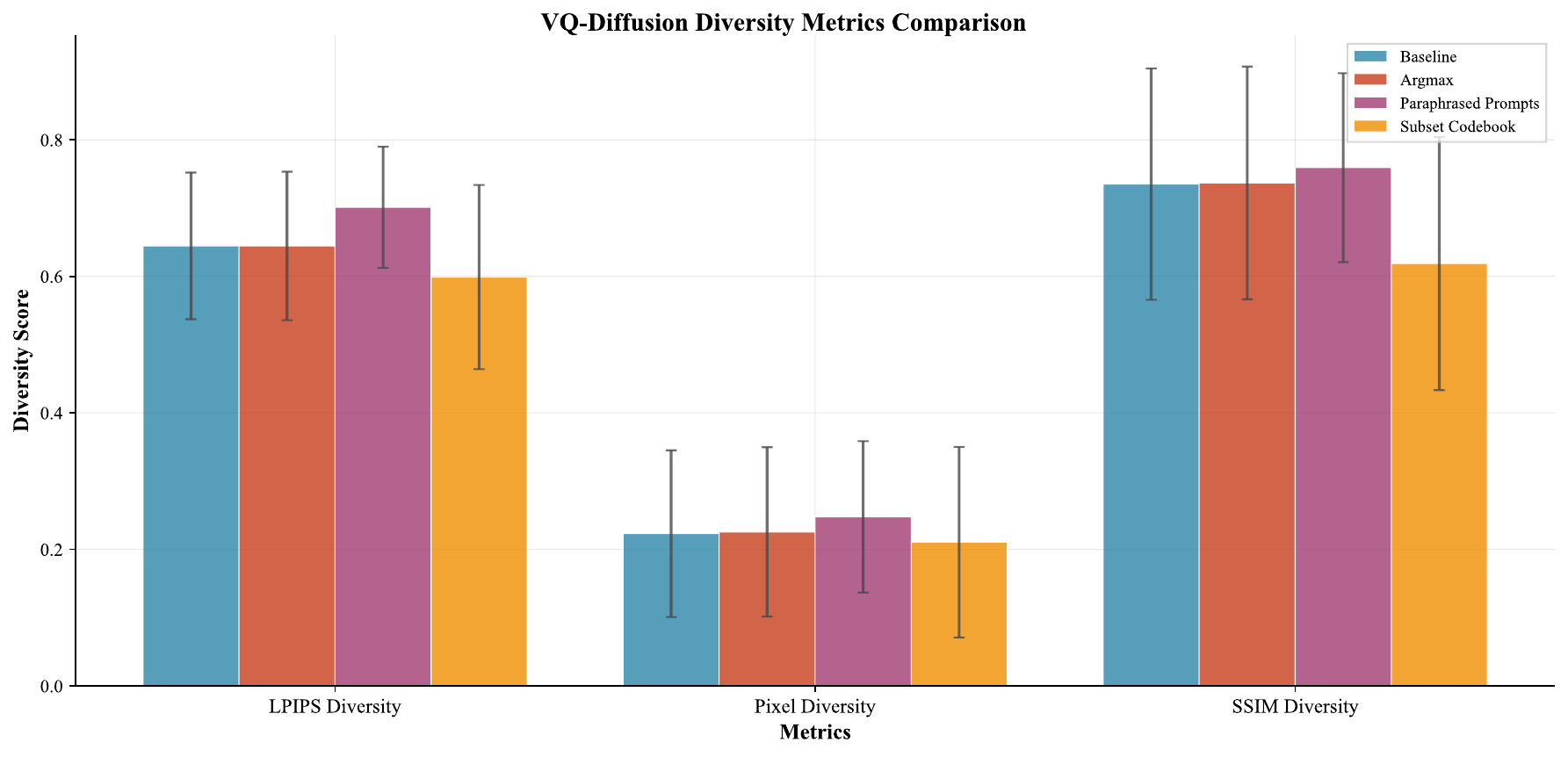}
        \caption{VQ-Diffusion Diversity Comparsion}
        \label{fig:vqdiff_disentangle}
    \end{subfigure}
    \hfill
    \begin{subfigure}[b]{0.33\textwidth}
        \centering
        \includegraphics[width=\linewidth]{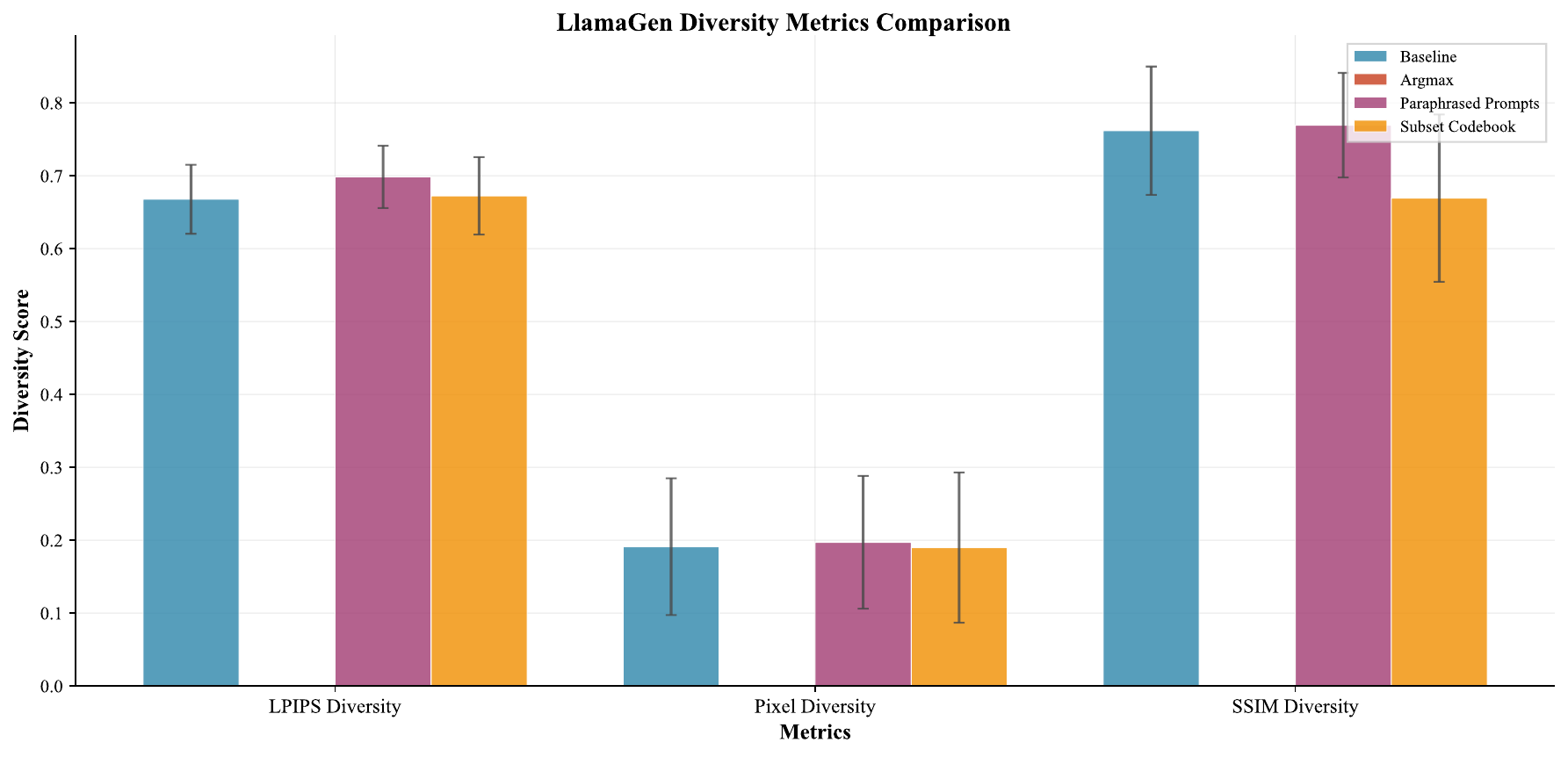}
        \caption{LlamaGen Diversity Comparsion}
        \label{fig:llamagen_disentangle}
    \end{subfigure}

    \caption{
Diversity analysis of three representative generative models under the proposed diagnostic framework. 
Each subfigure reports quantitative diversity metrics before and after the three inference-time interventions introduced in \cref{exp_probes}: 
the \textit{Codebook Subset} probe (measuring sensitivity to codebook entropy $H(Z)$), 
the \textit{Argmax} probe (isolating execution randomness $H_{\text{exec}}$), 
and the \textit{Paraphrase} probe (estimating conditional entropy $H(Z|X)$). 
(a) Masked Image Model (\textit{aMUSEd}), 
(b) Diffusion-based model (\textit{VQ-Diffusion}), 
and (c) Autoregressive model (\textit{LlamaGen}). 
Each bar corresponds to a distinct diversity metric, 
The comparison illustrates how different generative paradigms respond to compression and diversity pressures under controlled interventions. }

    \label{fig:disentangle_res}
\end{figure*}

%% file: figs/Figure5.tex
\begin{figure*}[tp]
    \centering
    \includegraphics[width=0.33\linewidth]{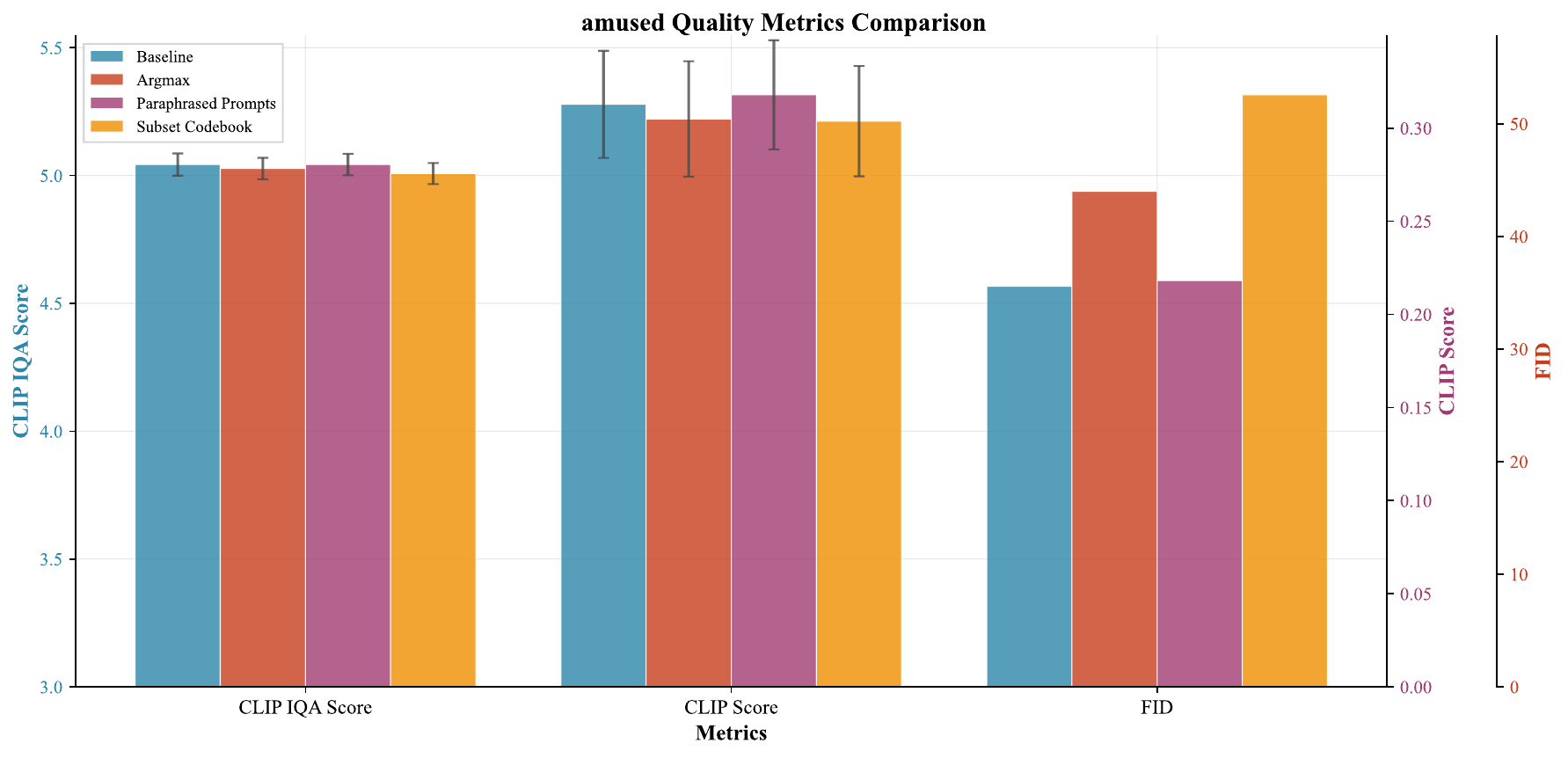}
    \hfill
    \includegraphics[width=0.33\linewidth]{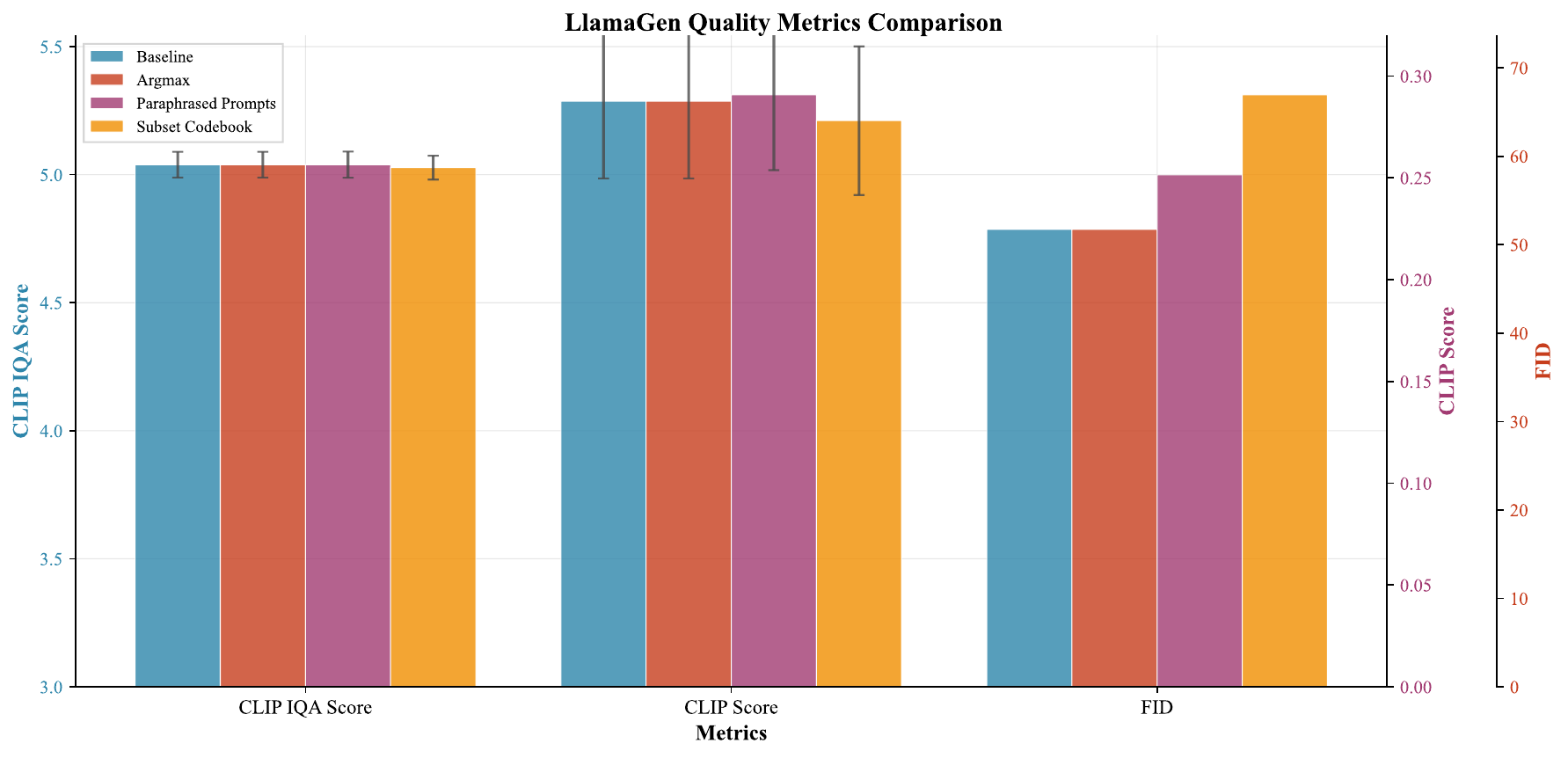}
    \hfill
    \includegraphics[width=0.33\linewidth]{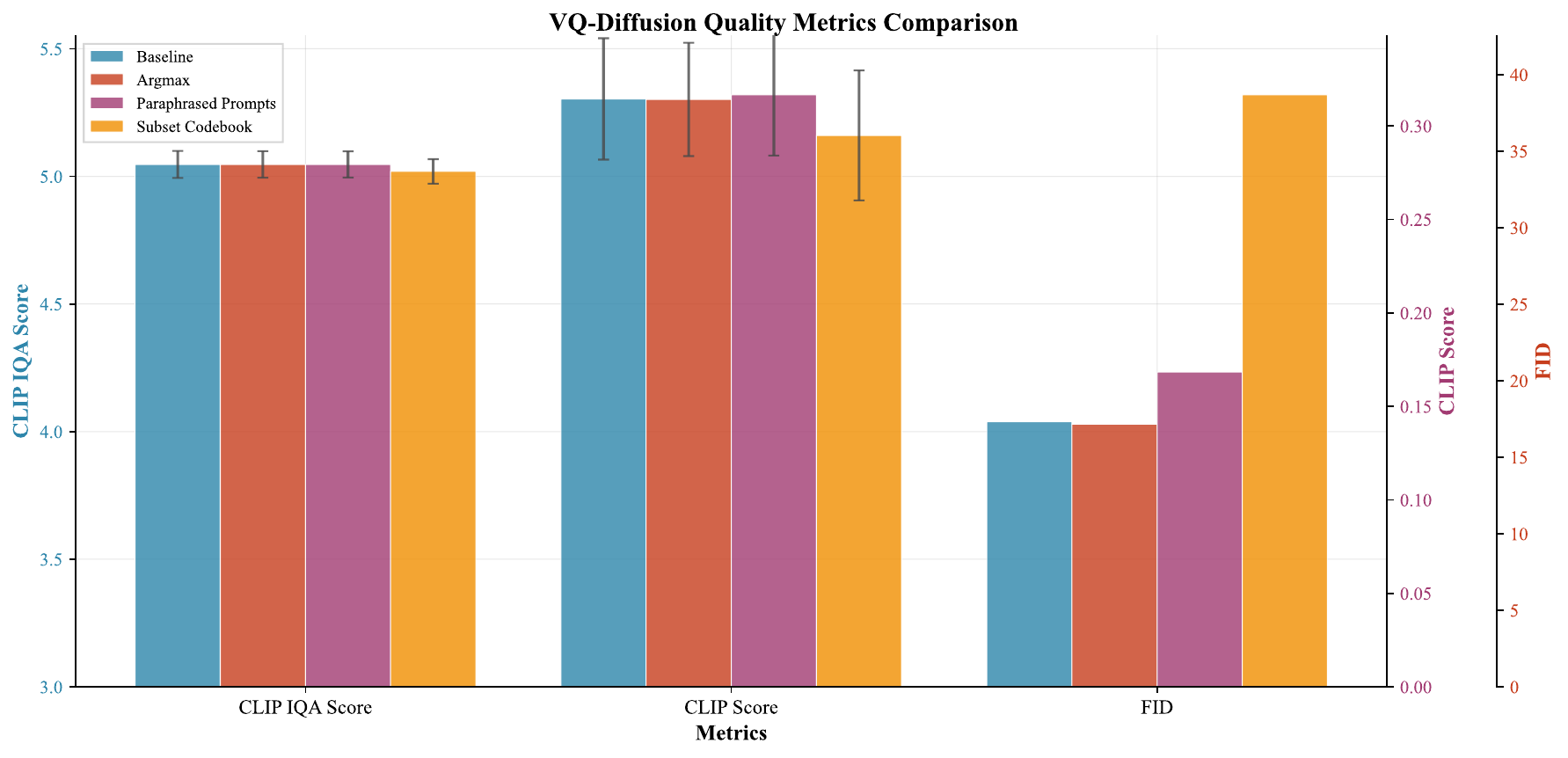}

    \caption{Impact of the three experimental probes on generation quality across all models. Quality is measured by CLIP Score (prompt-image alignment), CLIP IQA Score (image aesthetics) and FID (image fidelity). (Left) For aMUSEd, interventions that reduce diversity also degrade quality. (Center) For LlamaGen, interventions have minimal impact on quality, consistent with its collapsed state. (Right) For VQ-Diffusion, only the Subset intervention significantly impacts quality, revealing a decoupling of its diversity and quality mechanisms.}
    \label{fig:5}
\end{figure*}

%% file: figs/Figure6.tex
\begin{figure}[tp]
    \centering

    \begin{subfigure}[b]{0.48\columnwidth}
        \centering
        \includegraphics[width=\linewidth]{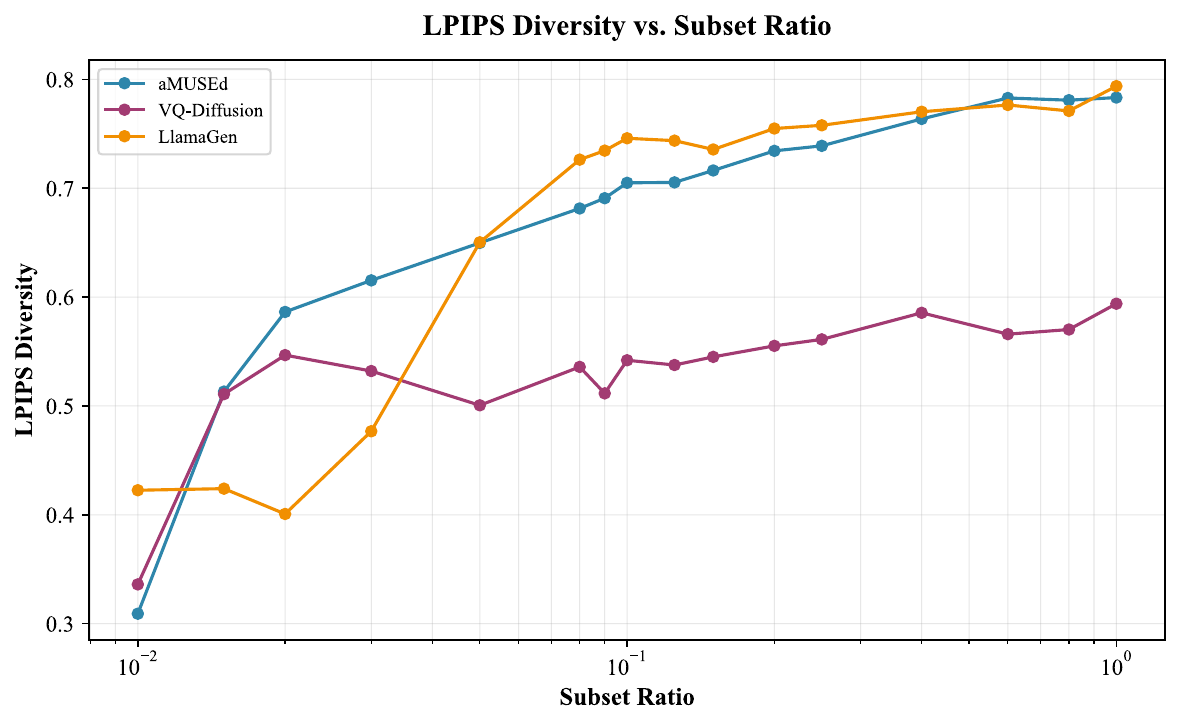}
        \caption{LPIPS Diversity vs. Subset Ratio}
        \label{fig:6L}
    \end{subfigure}
    \hfill 
    \begin{subfigure}[b]{0.48\columnwidth}
        \centering
        \includegraphics[width=\linewidth]{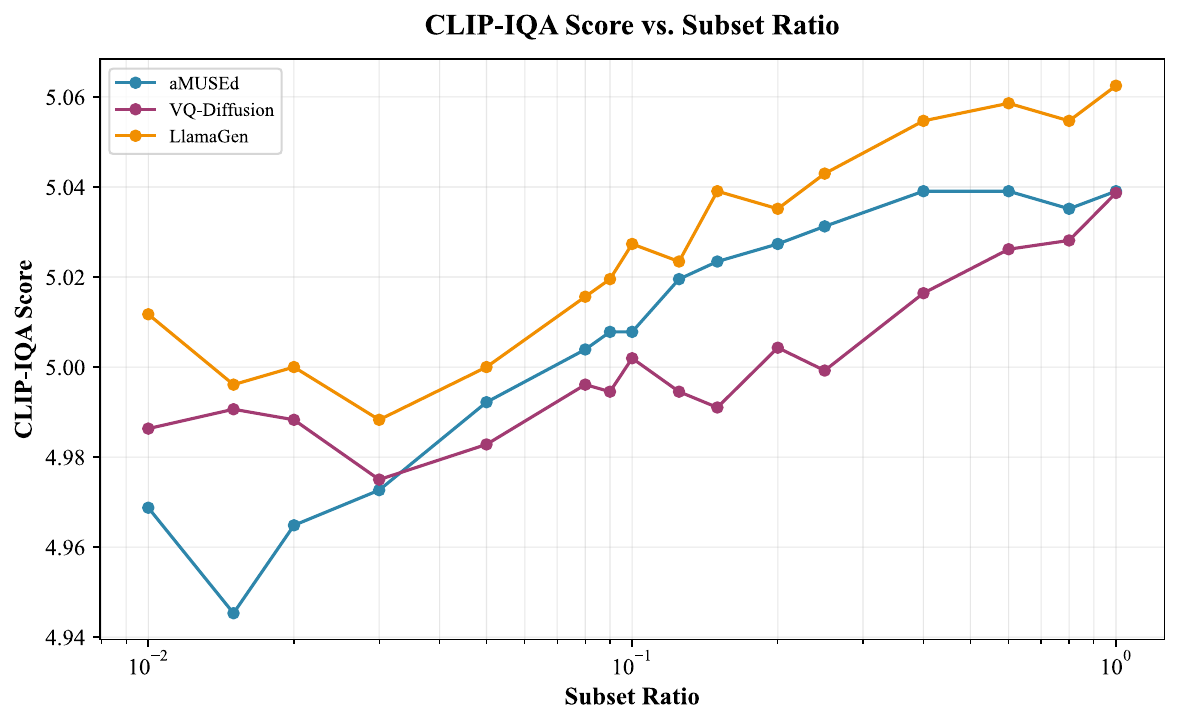}
        \caption{CLIP-IQA vs. Subset Ratio}
        \label{fig:6R}
    \end{subfigure}

    \caption{Ablation study on the codebook subset ratio for aMUSEd (MIM), VQ-Diffusion, and LlamaGen (AR). The curves reveal the different sensitivities of each model to codebook capacity, illustrating the fundamental trade-off between generative diversity and image quality. The Subset intervention in our main experiments was conducted at a ratio where diversity is highly sensitive but quality is not yet significantly compromised.}
    \label{fig:6}
\end{figure}

%% file: figs/ablations.tex
\begin{figure*}[tp]
    \centering

    \begin{subfigure}[b]{0.33\textwidth}
        \centering
        \includegraphics[width=\linewidth]{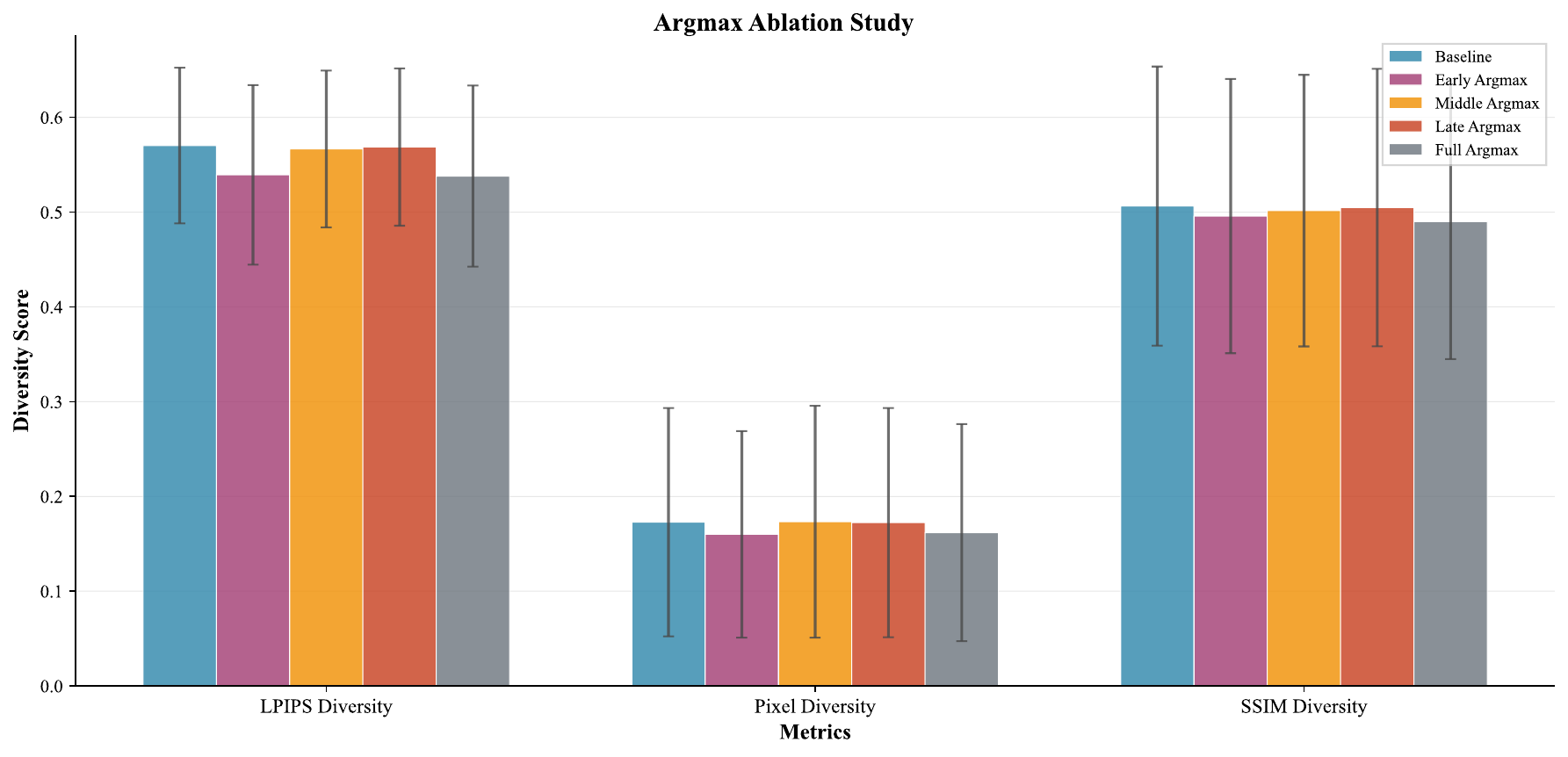}
        \caption{Argmax Timing Ablation}
        \label{fig:argmax_timing}
    \end{subfigure}
    \hfill 
    \begin{subfigure}[b]{0.33\textwidth}
        \centering
        \includegraphics[width=\linewidth]{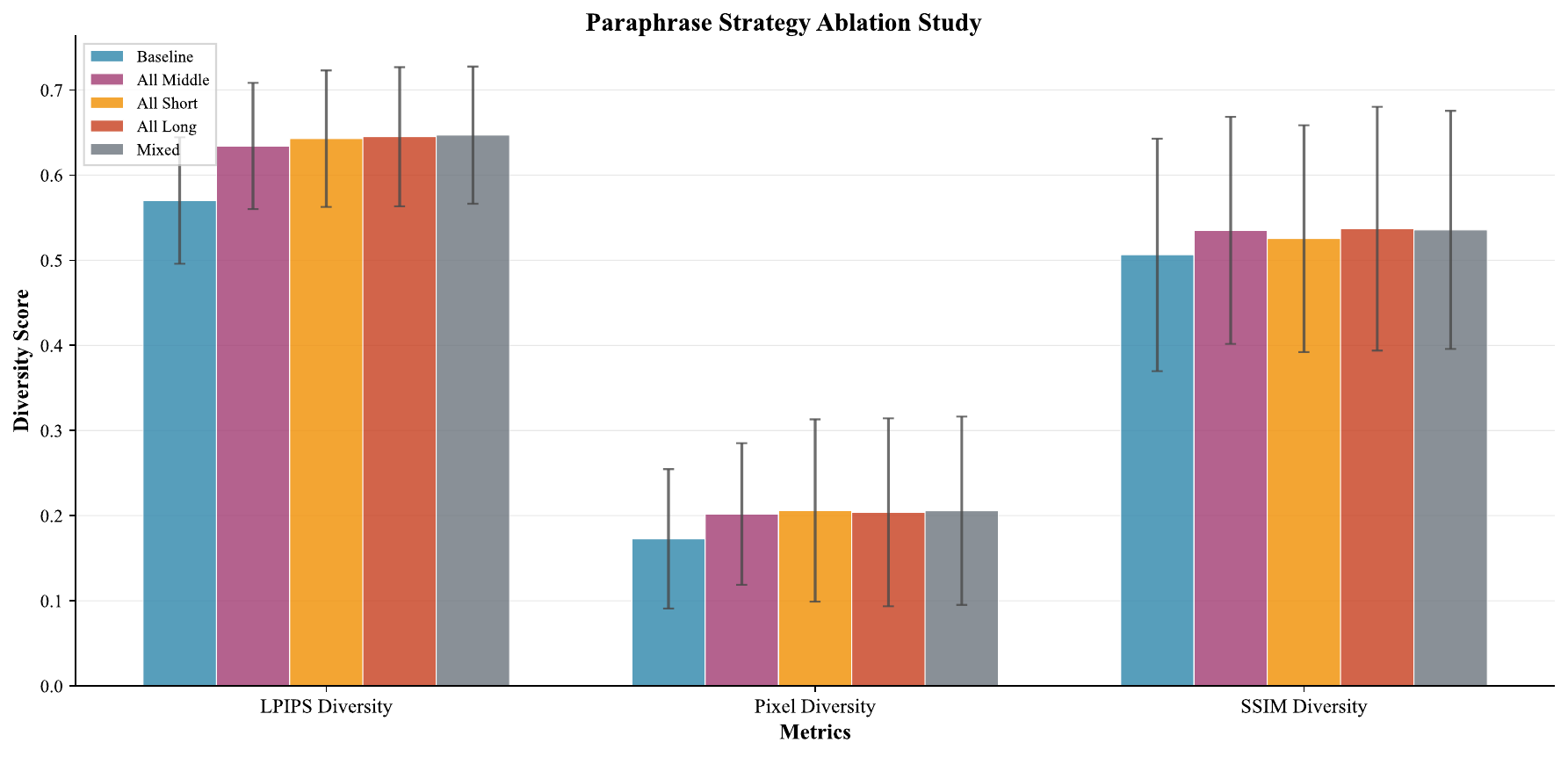}
        \caption{Paraphrase Length Ablation}
        \label{fig:prompt_length}
    \end{subfigure}
    \hfill
    \begin{subfigure}[b]{0.33\textwidth}
        \centering
        \includegraphics[width=\linewidth]{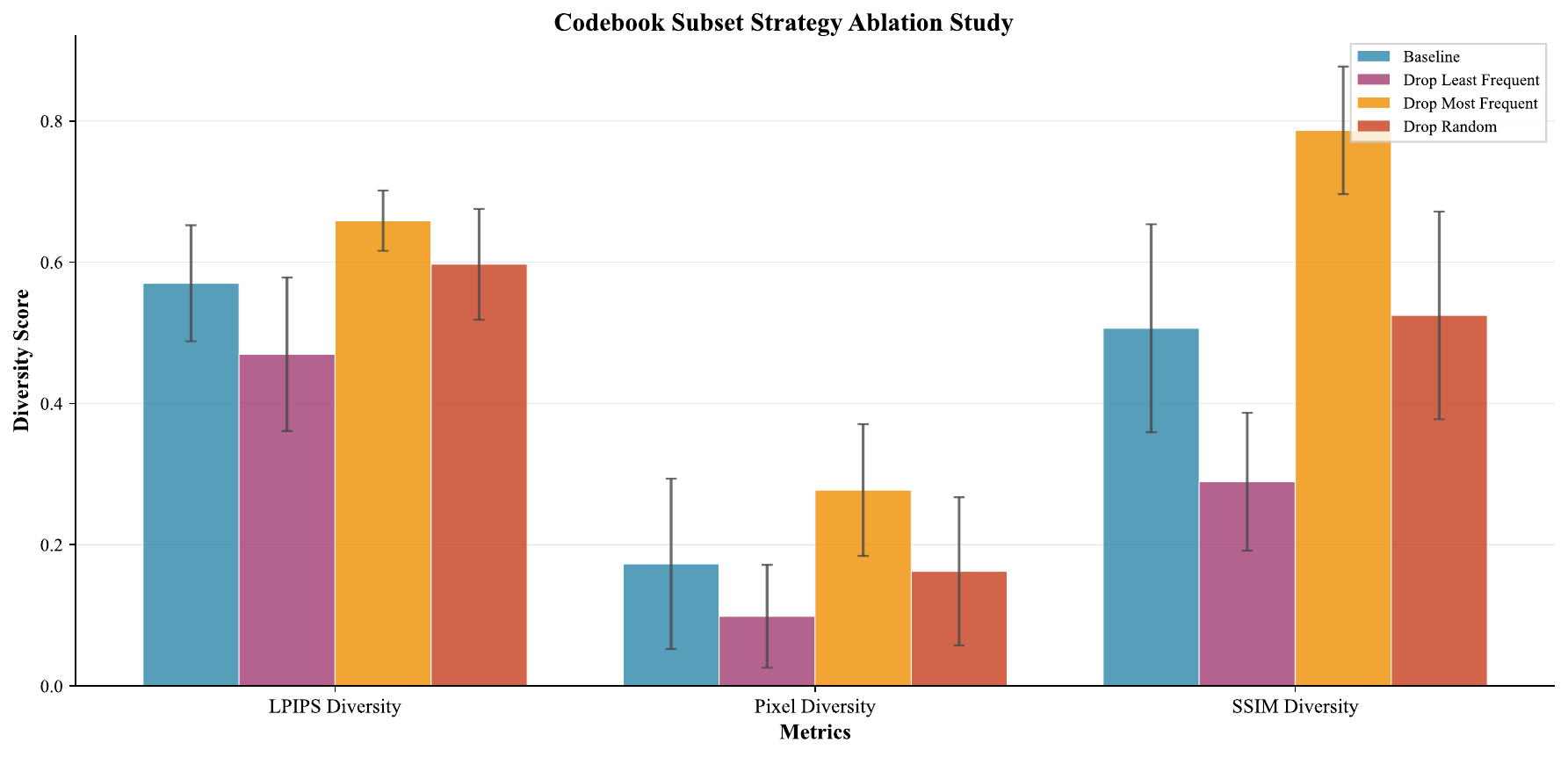}
        \caption{Codebook Subsetting Ablation}
        \label{fig:subset_method}
    \end{subfigure}
    
    \caption{Further ablation studies. (a) Impact on diversity when applying Argmax intervention during early, middle, or late stages of generation. (b) Impact on diversity using paraphrases of different lengths (short, middle, long) or a mixed set. (c) Impact on diversity when constructing the codebook subset by removing least frequent (original method), most frequent, or random codebook vectors.}
    \label{fig:further_ablations}
\end{figure*}

%% file: figs/VLM.tex
\begin{figure}[t!]
    \centering

    \begin{subfigure}[b]{0.48\columnwidth}
        \centering
        \includegraphics[width=\linewidth]{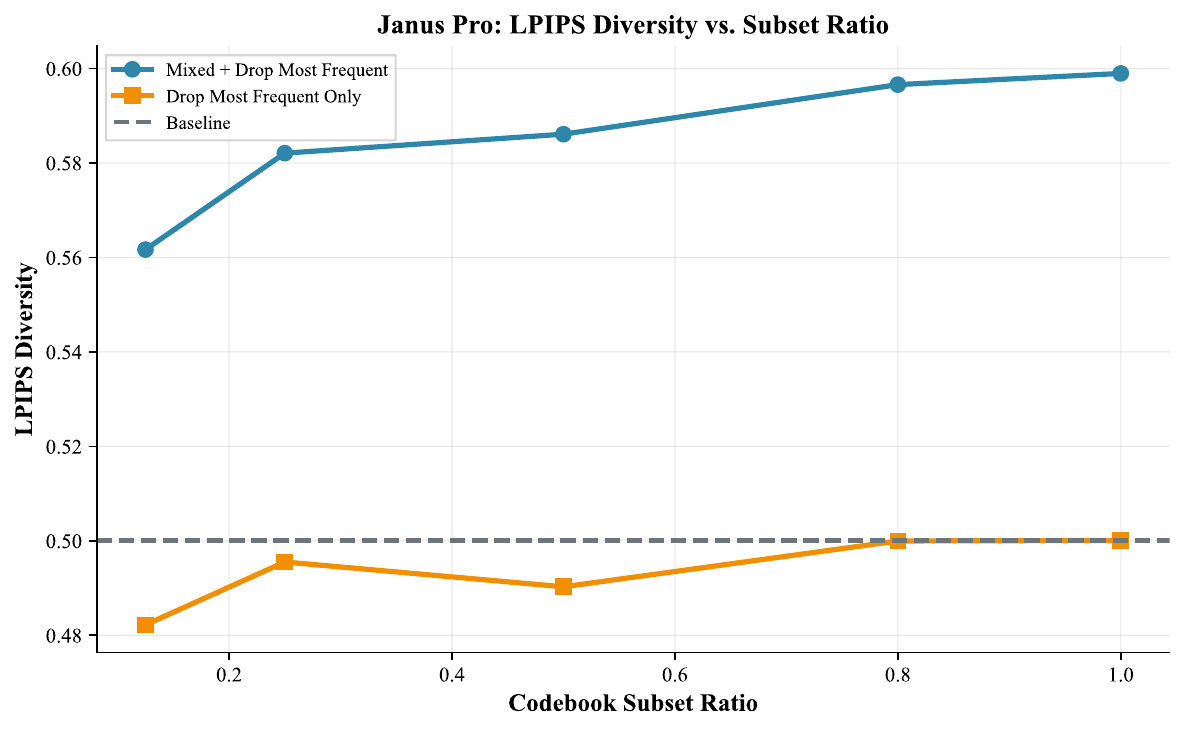}
        \caption{Janus: LPIPS Diversity vs. Subset Ratio}
        \label{fig:Janus_lpips_curve}
    \end{subfigure}
    \hfill 
    \begin{subfigure}[b]{0.48\columnwidth}
        \centering
        \includegraphics[width=\linewidth]{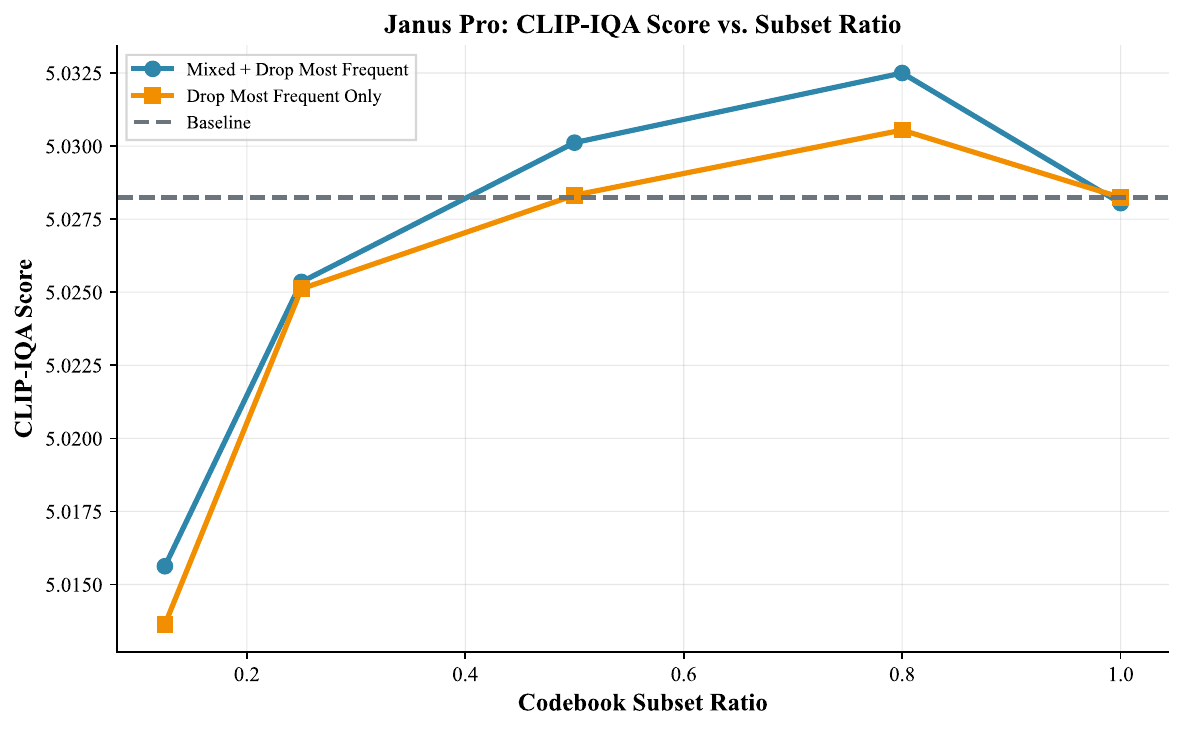}
        \caption{Janus: CLIP-IQA Score vs. Subset Ratio}
        \label{fig:Janus_iqa_curve}
    \end{subfigure}

    
    \begin{subfigure}[b]{0.48\columnwidth}
        \centering
        \includegraphics[width=\linewidth]{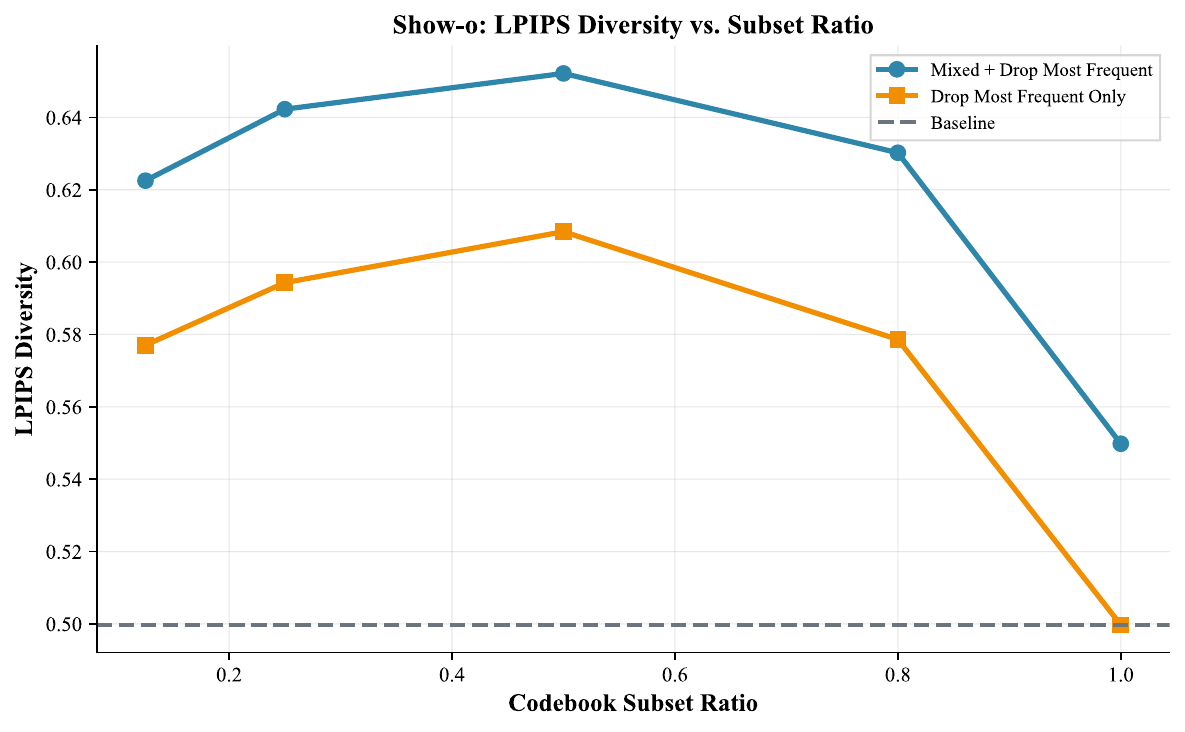}
        \caption{Show-O: LPIPS Diversity vs. Subset Ratio}
        \label{fig:show-o_lpips_curve}
    \end{subfigure}
    \hfill 
    \begin{subfigure}[b]{0.48\columnwidth}
        \centering
        \includegraphics[width=\linewidth]{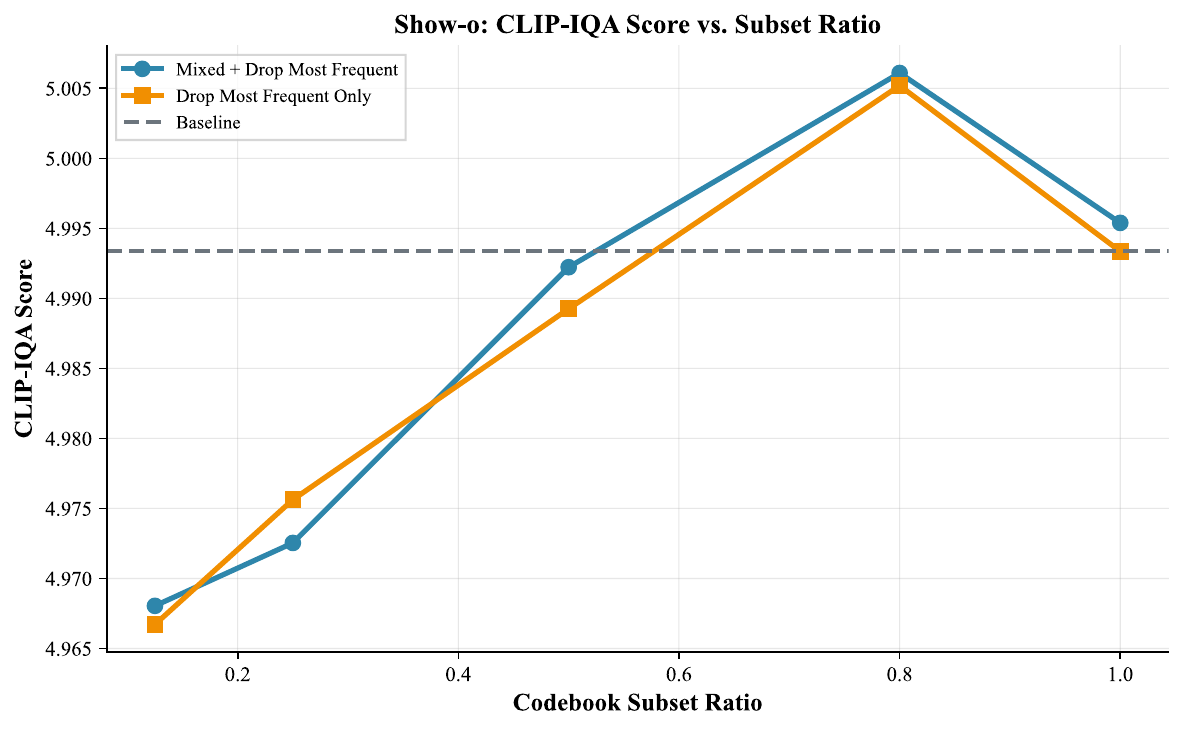}
        \caption{Show-O: CLIP-IQA Score vs. Subset Ratio}
        \label{fig:show-o_iqa_curve}
    \end{subfigure}
    
    \caption{Preliminary validation of the proposed diversity enhancement strategy on DeepSeek Janus Pro 1B (top row) and VLM Show-O (bottom row). The strategy combines mixed paraphrases with disabling the most frequent codebook tokens. (a, c) LPIPS diversity consistently increases for both models compared to their respective baselines. (b, d) CLIP-IQA score shows a trade-off, but remains competitive, peaking at certain subset ratios. }
    \label{fig:vlm_diversity_enhancement}
\end{figure}

%% file: figs/samples.tex
\begin{figure}[htbp]
    \centering
    \newcommand{\imgwidth}{0.2\columnwidth} 

    \begin{tabular}{c c c c}
        \textbf{Baseline} & 
        \textbf{Subset} & 
        \textbf{Paraphrase} & 
        \textbf{Proposed}\\
        \toprule
        
        \includegraphics[width=\imgwidth]{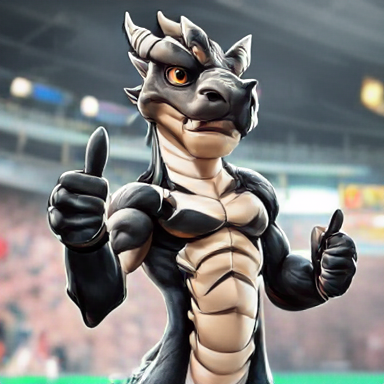} &
        \includegraphics[width=\imgwidth]{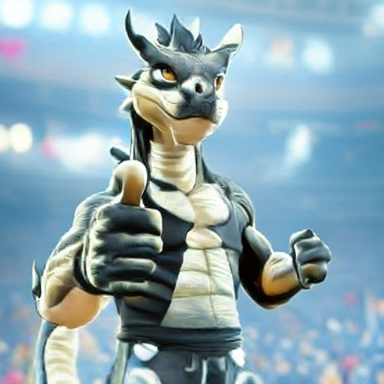} &
        \includegraphics[width=\imgwidth]{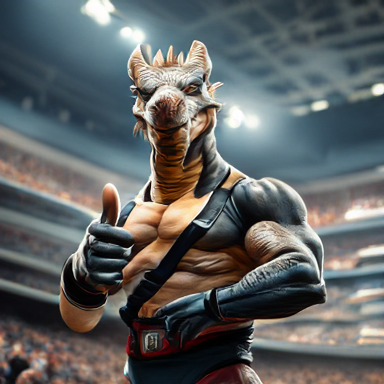} &
        \includegraphics[width=\imgwidth]{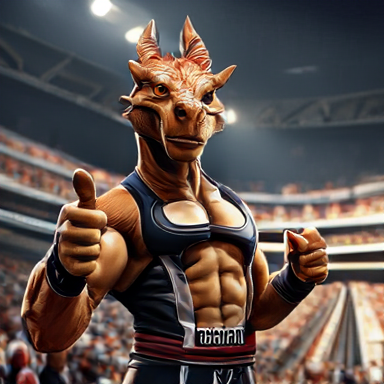} \\
                
        \includegraphics[width=\imgwidth]{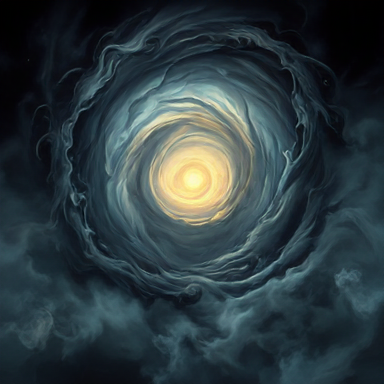} &
        \includegraphics[width=\imgwidth]{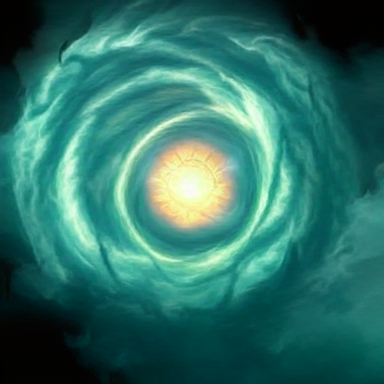} &
        \includegraphics[width=\imgwidth]{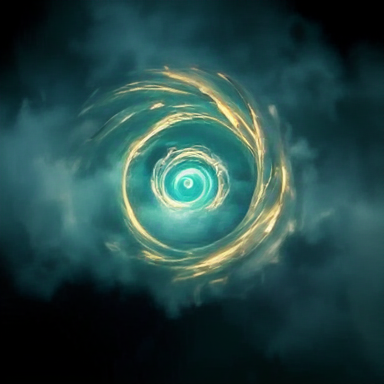} &
        \includegraphics[width=\imgwidth]{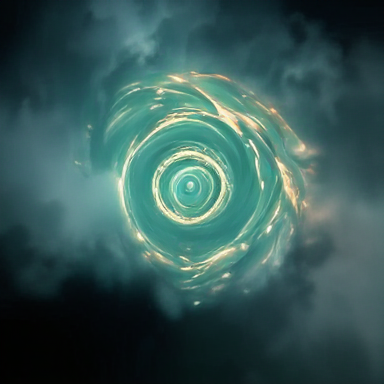} \\
        
        \includegraphics[width=\imgwidth]{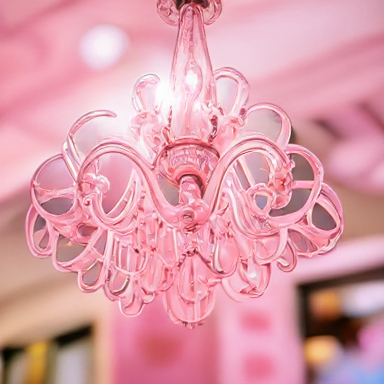} &
        \includegraphics[width=\imgwidth]{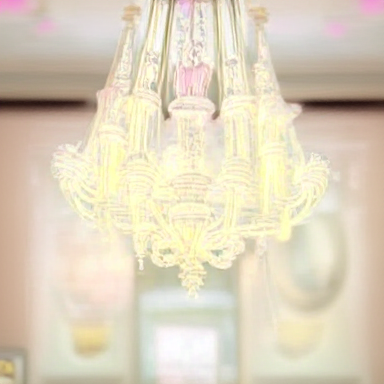} &
        \includegraphics[width=\imgwidth]{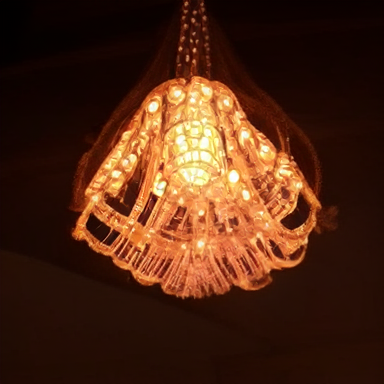} &
        \includegraphics[width=\imgwidth]{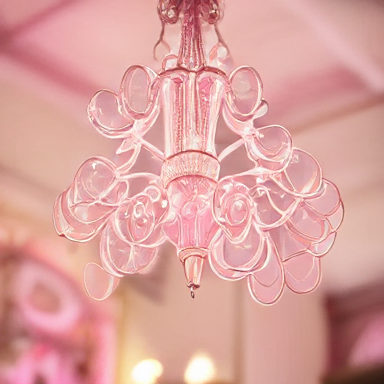} \\
        
        \bottomrule
    \end{tabular}
    
    \caption{Qualitative comparison of generative samples under different inference strategies. Each \textbf{column} represents a fixed strategy, while each \textbf{row} shows a different generative sample (e.g., from a different base prompt). This grid illustrates the consistent visual effect of each intervention across multiple examples generated by Janus Pro. The 'Proposed' column represents using mixed paraphrased prompts and drop most frequent 20\% tokens}
    \label{fig:qualitative_grid}
\end{figure}

%% file: 10_conclusion.tex
\section{Conclusion}
\label{sec:conclusion}

We presented an information-theoretic framework for analyzing generative diversity in discrete latent models. By formulating the generation process as an Information Bottleneck trade-off between compression and diversity, and decomposing the latter into path- and execution-level sources, our approach connects empirical diversity patterns to interpretable quantities. Through controlled inference-time interventions, we show that AR, MIM, and diffusion models embody distinct resolutions of this trade-off, each reflecting a characteristic balance between entropy and fidelity.

Beyond offering theoretical insight, our framework enables practical diagnostics for pretrained generative systems and guides diversity modulation without retraining. The broader implication is that generative diversity can be measured, reasoned about, and systematically controlled through the lens of information theory—paving the way for more predictable and tunable creative behavior in large-scale generative models.

%% file: 12_appendix.tex
\section{Appendix Section}
\label{sec:appendix_section}
\subsection{Analysis of Generative Factor Interactions}
\label{sec:interactions}

\input{figs/amused_waterfall}
\input{figs/vqdiff_waterfall}
\input{figs/llamagen_watefall}

\input{figs/amused_intersection}
\input{figs/vqdiff_intersection}
\input{figs/llamagen_intersection}

To further delineate the models' strategies, we analyzed the factorial interactions between Sampling Strategy ($H_{exec}$), Codebook State ($H(Z)$), and Prompt Type ($H(Z|X)$). This reveals deeper, non-additive relationships masked by the analysis of main effects alone.

For \textbf{aMUSEd}, the waterfall analysis (\cref{fig:waterfall_amused}) shows a negligible net three-way interaction ("Synergy Gap" of +0.001). However, the interaction profiles (\cref{fig:interactions_amused}) reveal this near-additivity is a macro-level artifact resulting from strong, opposing interactions. Specifically, a significant \textbf{crossover interaction} exists between sampling strategy and prompt type: deterministic sampling reduces diversity for original prompts but increases it for paraphrased prompts. This demonstrates a potent \textbf{compensatory mechanism} where the model amplifies sensitivity to $H(Z|X)$ when $H_{exec}$ is constrained. This effect is further magnified when $H(Z)$ is also limited (subset codebook), indicating a dynamic strategy that actively shifts reliance between mechanisms to maintain diversity under varying constraints.

In stark contrast, \textbf{VQ-Diffusion} exhibits genuine additivity. Its waterfall analysis (\cref{fig:waterfall_vqdiff}) shows a synergy gap of precisely zero. The interaction profiles (\cref{fig:interactions_vqdiff}) confirm this, displaying nearly parallel lines across all facets. The minimal impact of the Argmax intervention ($H_{exec}$) and the negligible effect of paraphrasing ($H(Z|X)$) remain consistent regardless of other factors. This lack of significant interaction supports the "Decoupled Strategy," where path diversity ($H_{path}$) is the primary driver, operating largely independently from execution stochasticity, input sensitivity, and codebook utilization.

\textbf{LlamaGen}, whose diversity relies entirely on $H_{exec}$, requires a two-factor analysis (Prompt Type $\times$ Codebook State). Its waterfall analysis (\cref{fig:waterfall_llamagen}) yields a zero synergy gap, suggesting additivity. This is strongly supported by the interaction profile in \cref{fig:interaction_llamagen}. The lines representing the Full and Subset codebook states are \textbf{nearly parallel}, indicating a \textbf{negligible interaction} between prompt type and codebook state. While paraphrasing consistently increases diversity (confirming sensitivity to $H(Z|X)$), the magnitude of this increase is largely independent of whether the full or subset codebook is used. This near-additivity aligns with the "Compress-Prioritized" strategy, where the model operates within a constrained space, and the effects of input variation and limited codebook access do not significantly interfere with each other.

%% file: figs/amused_waterfall.tex
\begin{figure}[tp]
    \centering
    \includegraphics[width=\linewidth]{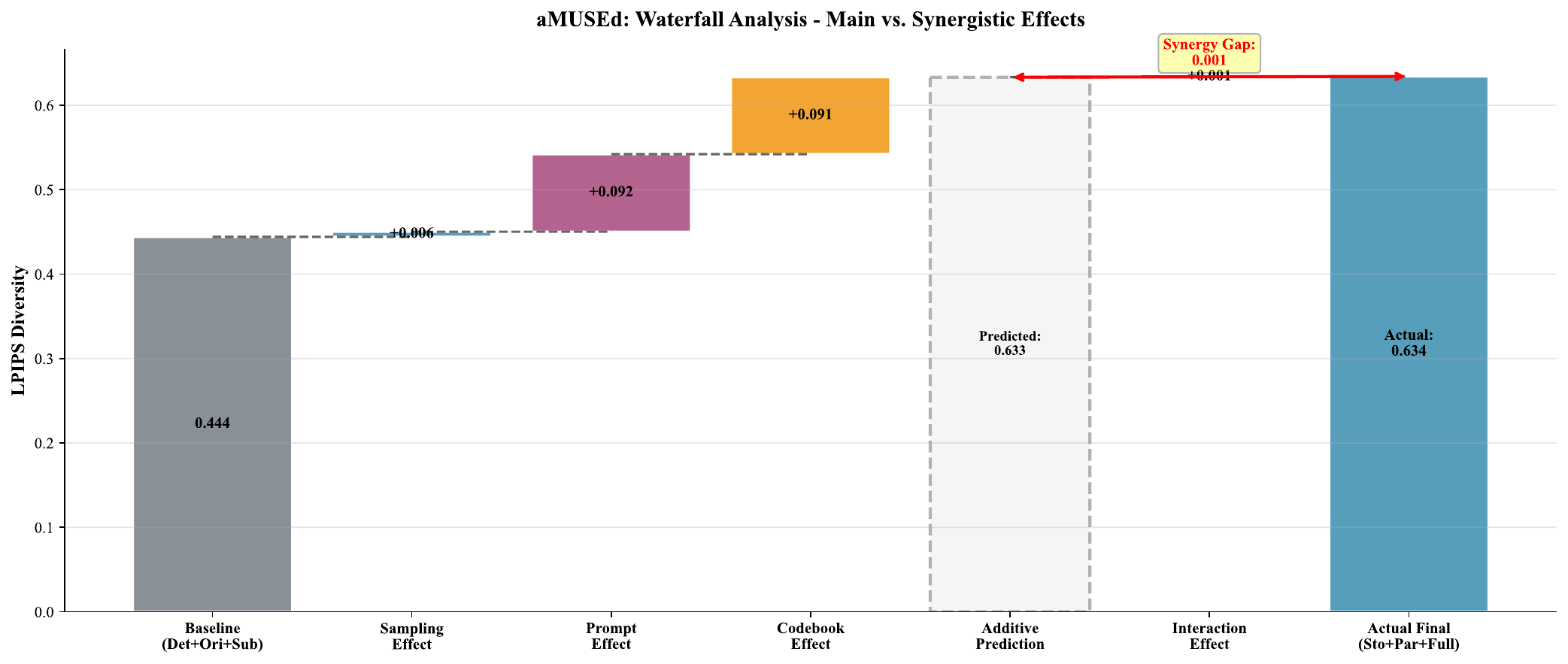}
    \caption{Waterfall analysis for aMUSEd. Starting from the baseline condition (Deterministic, Original, Subset), we cumulatively add the main effect of each generative factor. The final "Additive Prediction" (0.633) is the theoretical sum of these independent effects. This is compared to the "Actual Final" diversity (0.634) measured under the optimal condition (Stochastic, Paraphrased, Full). The "Synergy Gap" (+0.001) represents the net three-way interaction effect, which in this macro view appears negligible.}
    \label{fig:waterfall_amused}
\end{figure}

%% file: figs/vqdiff_waterfall.tex
\begin{figure}[tp]
    \centering
    \includegraphics[width=\linewidth]{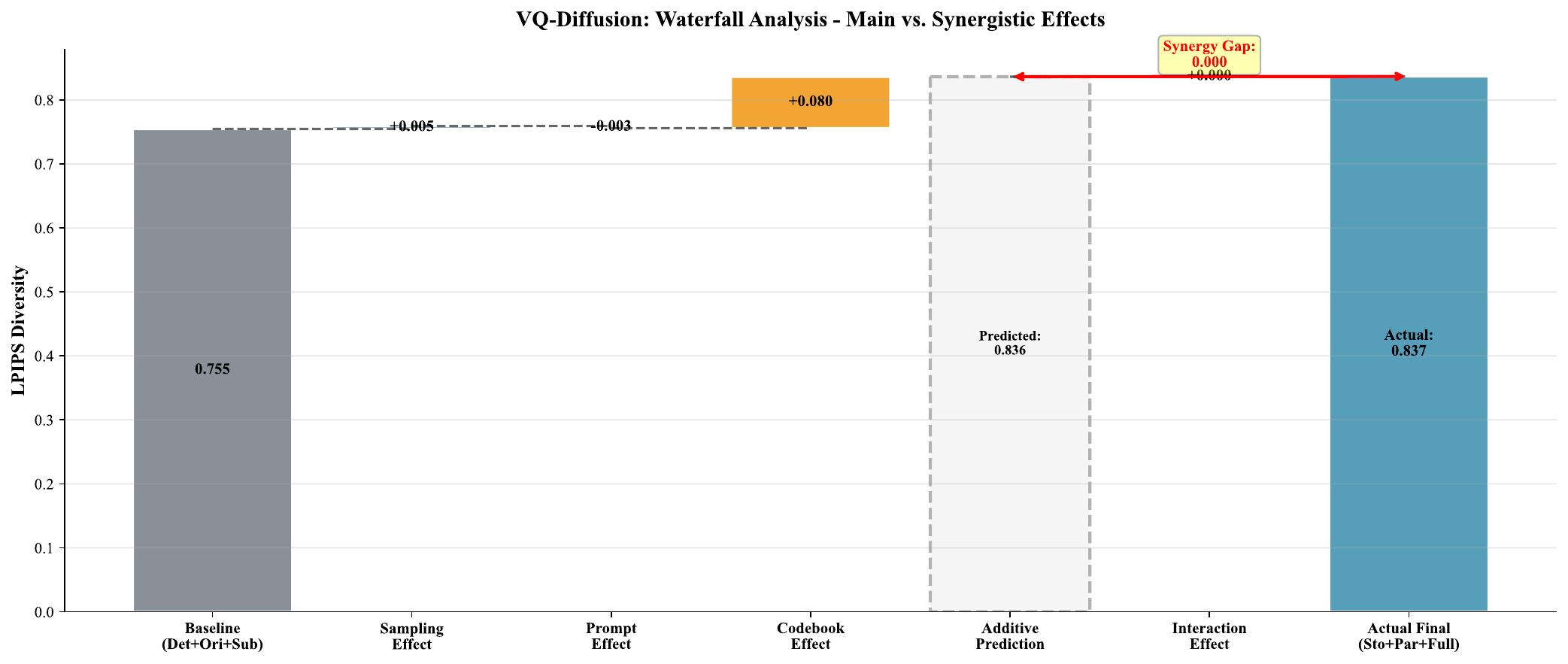}
    \caption{Waterfall analysis for VQ-Diffusion. Starting from the baseline condition (Deterministic, Original, Subset) with a diversity of 0.755, we cumulatively add the main effect of stochastic sampling (+0.005), paraphrased prompts (-0.003), and the full codebook (+0.080). The "Additive Prediction" (0.837) exactly matches the "Actual Final" diversity measured under the optimal condition (Stochastic, Paraphrased, Full), resulting in a zero "Synergy Gap".}
    \label{fig:waterfall_vqdiff}
\end{figure}

%% file: figs/llamagen_watefall.tex
\begin{figure}[tp]
    \centering
    \includegraphics[width=\linewidth]{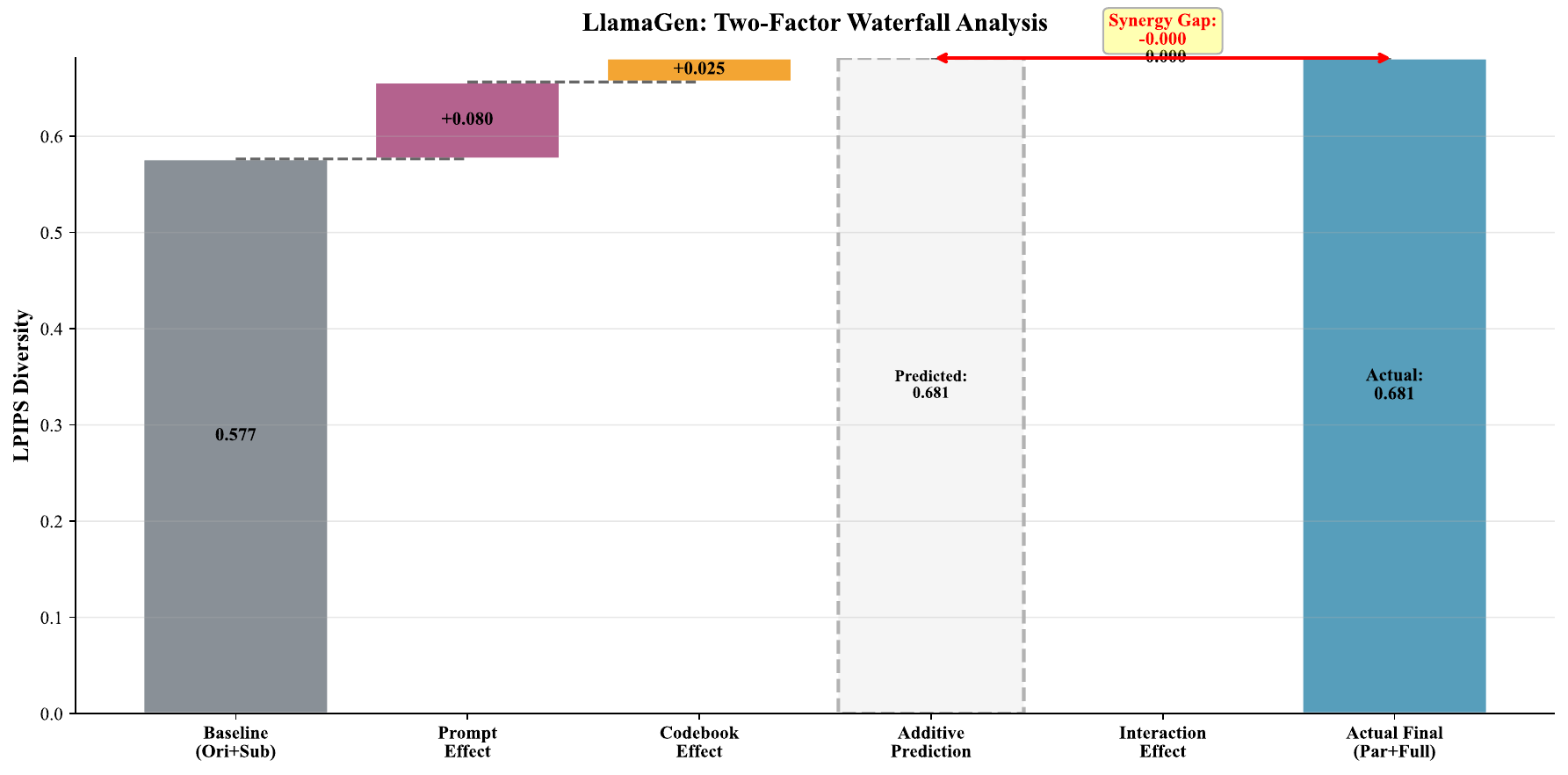}
    \caption{Two-factor waterfall analysis for LlamaGen (stochastic sampling implied). Starting from the baseline condition (Original prompt, Subset codebook) with a diversity of 0.577, we cumulatively add the main effect of paraphrased prompts (+0.080) and the full codebook (+0.025). The "Additive Prediction" (0.681) exactly matches the "Actual Final" diversity (Paraphrased, Full codebook), resulting in a zero "Synergy Gap". Note: Sampling effect is excluded as Argmax yields zero diversity.}
    \label{fig:waterfall_llamagen}
\end{figure}

%% file: figs/amused_intersection.tex
\begin{figure*}[tp]
    \centering

    \begin{subfigure}[b]{0.32\textwidth}
        \centering
        \includegraphics[width=\linewidth]{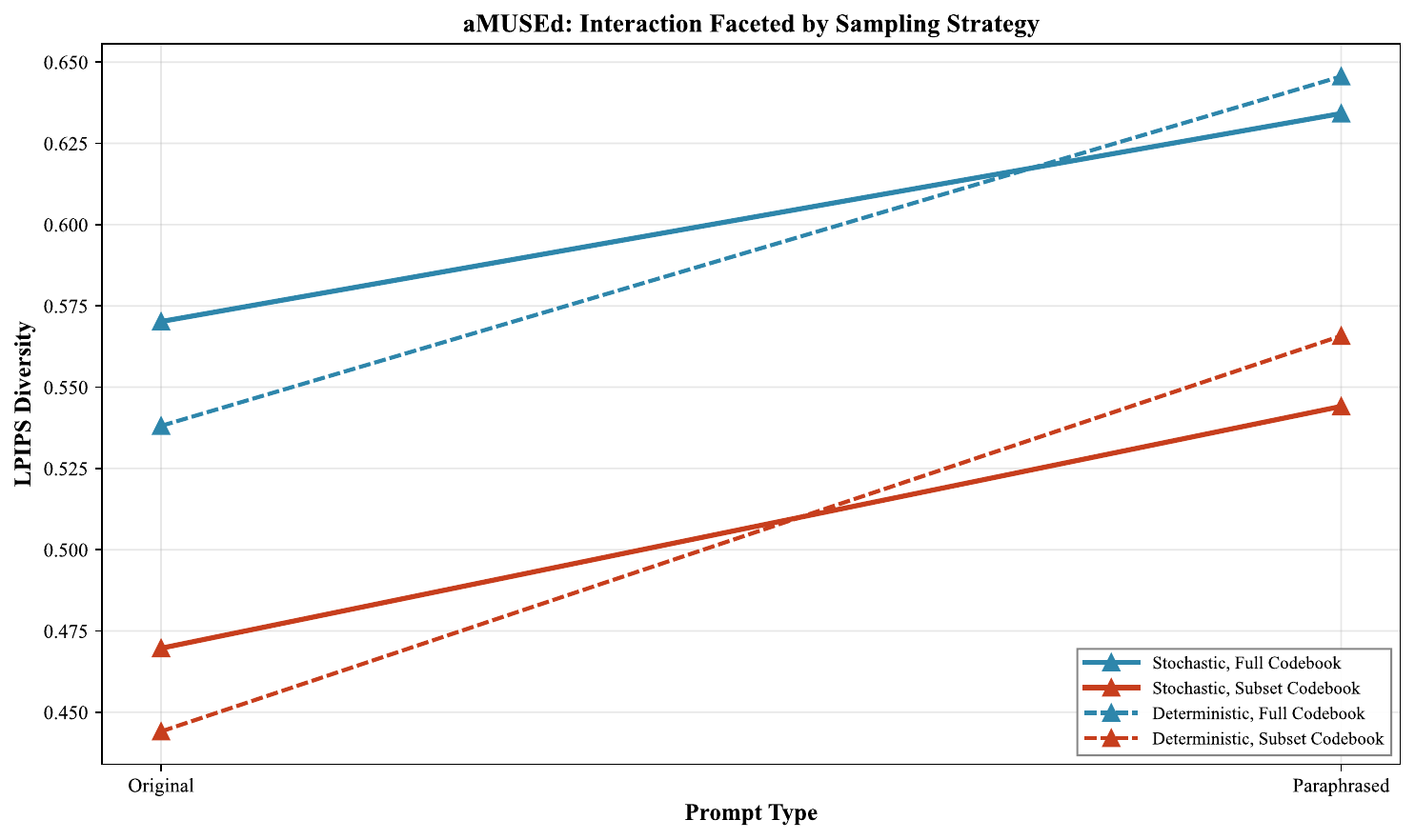}
        \caption{Faceted by Sampling Strategy}
        \label{fig:inter_amused_facet_sampling}
    \end{subfigure}
    \hfill 
    \begin{subfigure}[b]{0.32\textwidth}
        \centering
        \includegraphics[width=\linewidth]{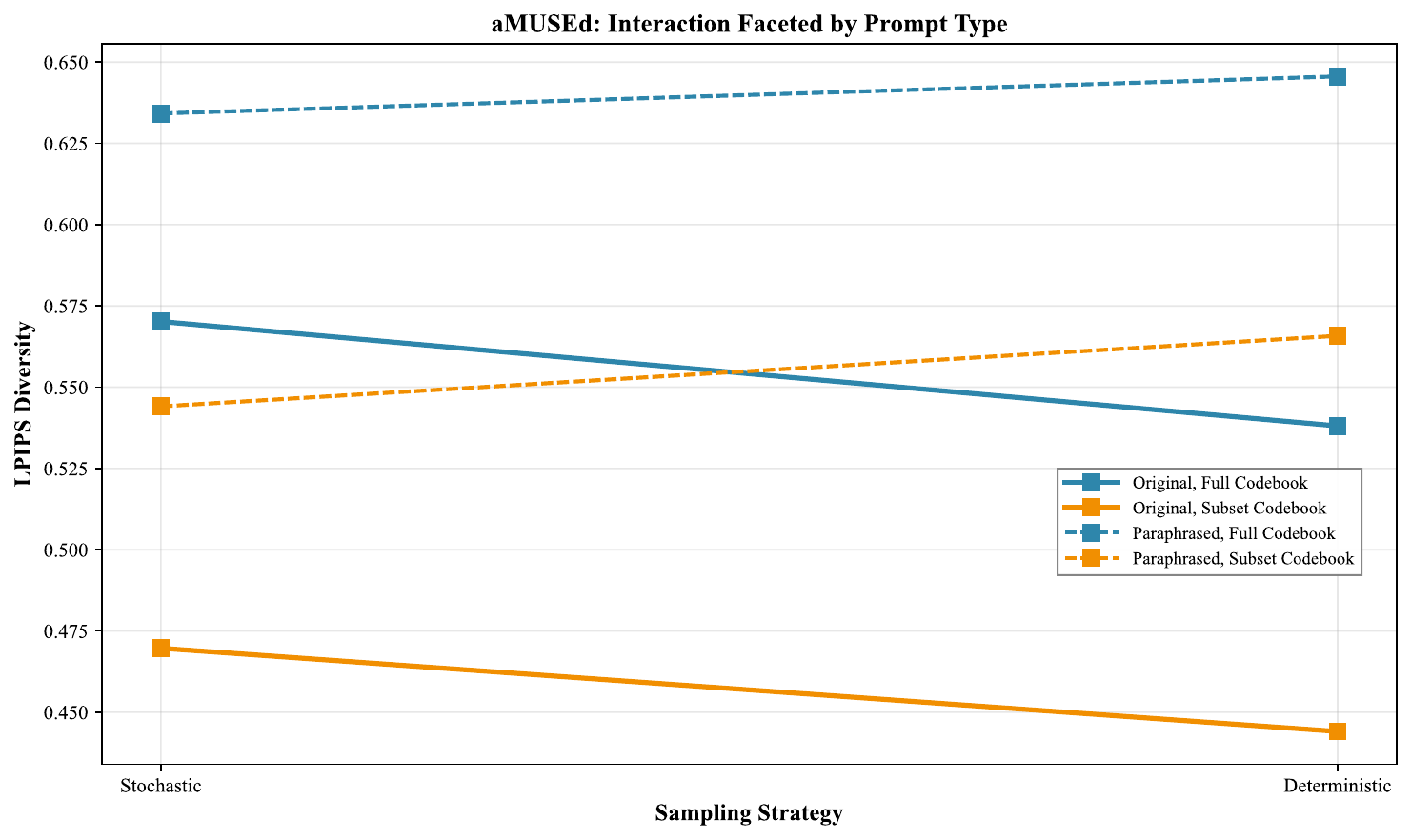}
        \caption{Faceted by Prompt Type}
        \label{fig:inter_amused_facet_prompt}
    \end{subfigure}
    \hfill
    \begin{subfigure}[b]{0.32\textwidth}
        \centering
        \includegraphics[width=\linewidth]{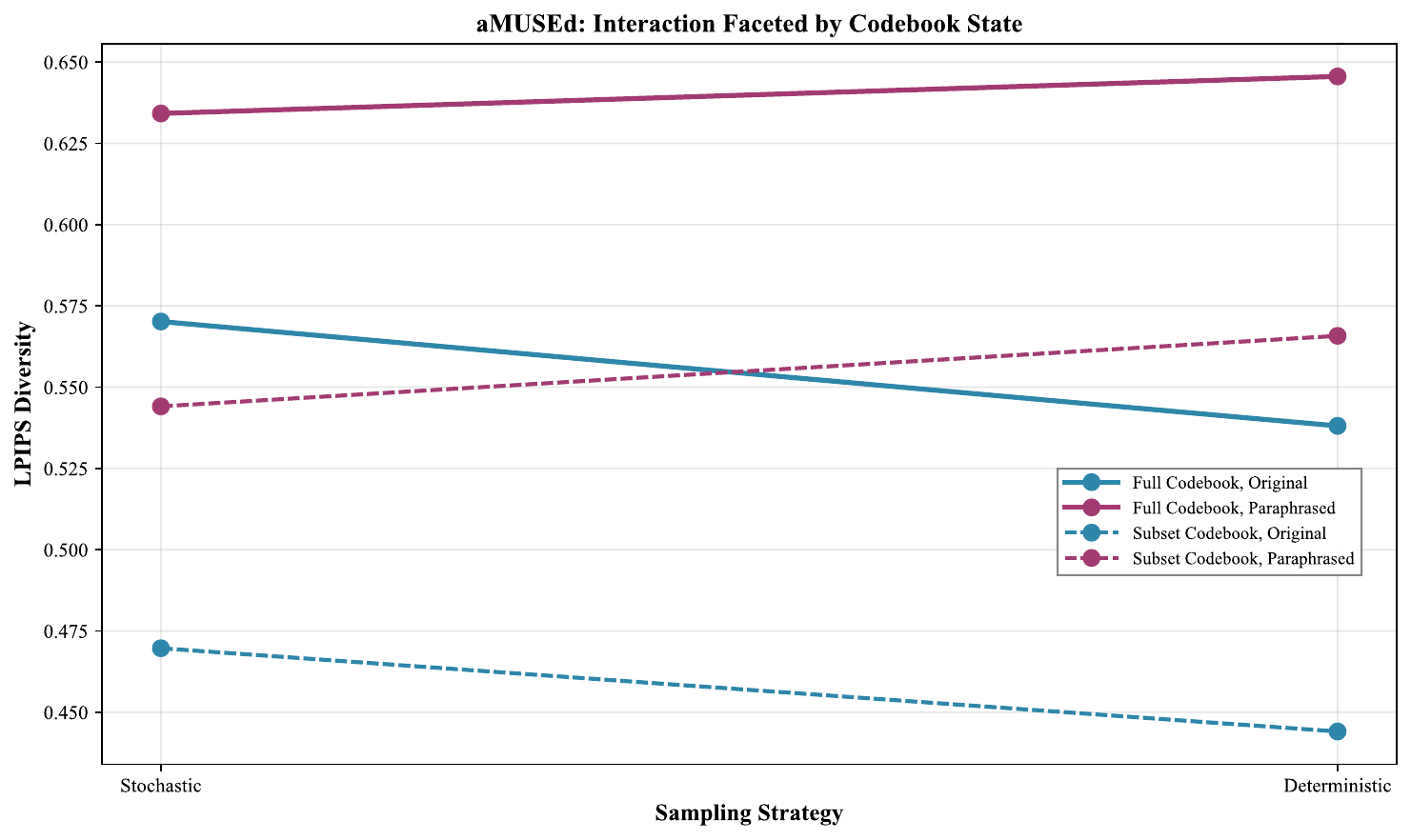}
        \caption{Faceted by Codebook State}
        \label{fig:inter_amused_facet_codebook}
    \end{subfigure}

    \caption{Three-way interaction profiles for aMUSEd, plotting LPIPS diversity across all 8 experimental conditions. (a) Faceting by Sampling Strategy shows that the diversity gain from paraphrasing (slope) is steeper under the Deterministic condition. (b, c) Faceting by Prompt Type and Codebook State reveals a strong crossover interaction: deterministic sampling \textit{decreases} diversity for Original prompts but \textit{increases} it for Paraphrased prompts, demonstrating a clear compensatory mechanism.}
    \label{fig:interactions_amused}
\end{figure*}

%% file: figs/vqdiff_intersection.tex
\begin{figure*}[tp]
    \centering

    \begin{subfigure}[b]{0.32\textwidth}
        \centering
        \includegraphics[width=\linewidth]{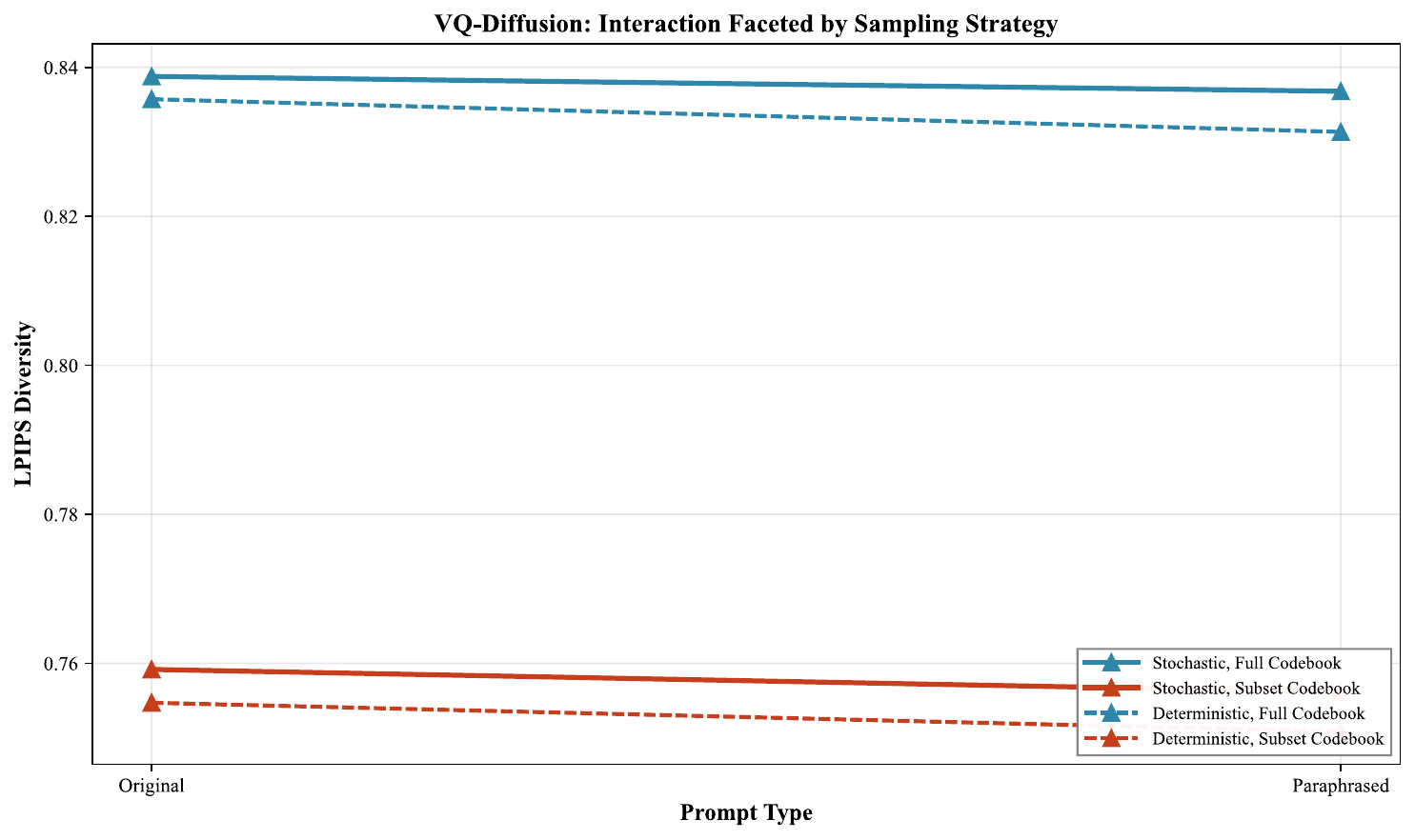}
        \caption{Faceted by Sampling Strategy}
        \label{fig:inter_vqdiff_facet_sampling}
    \end{subfigure}
    \hfill 
    \begin{subfigure}[b]{0.32\textwidth}
        \centering
        \includegraphics[width=\linewidth]{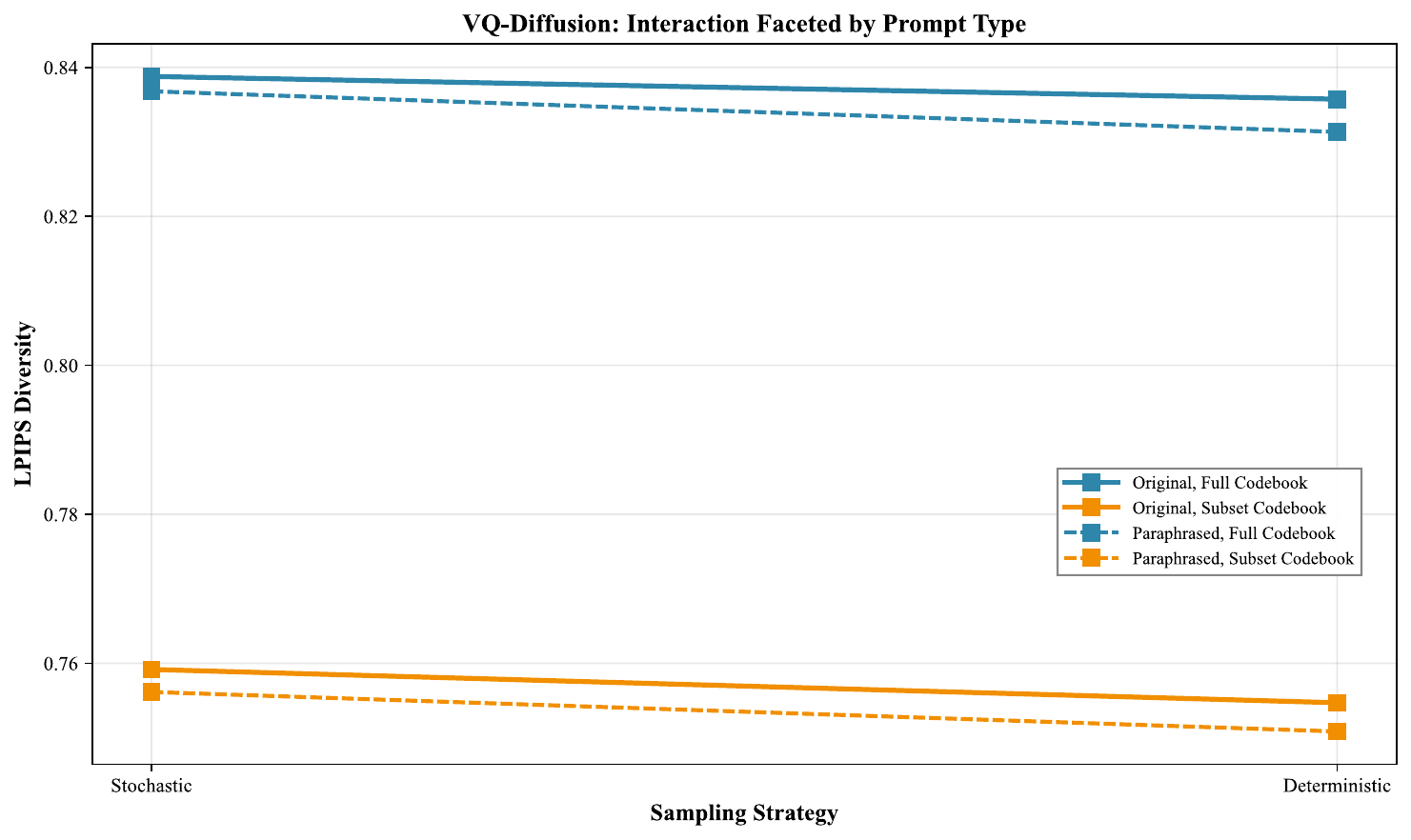}
        \caption{Faceted by Prompt Type}
        \label{fig:inter_vqdiff_facet_prompt}
    \end{subfigure}
    \hfill
    \begin{subfigure}[b]{0.32\textwidth}
        \centering
        \includegraphics[width=\linewidth]{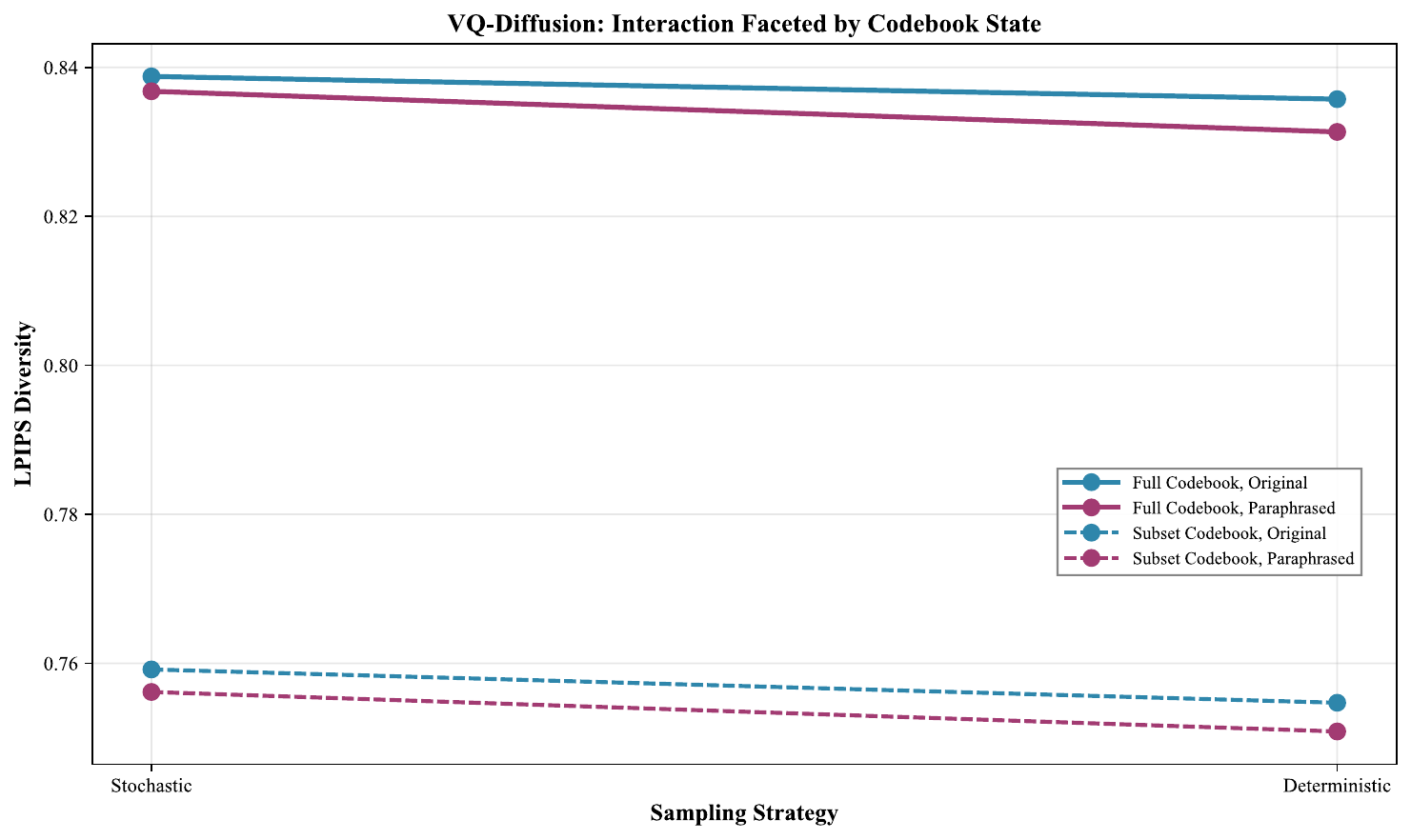}
        \caption{Faceted by Codebook State}
        \label{fig:inter_vqdiff_facet_codebook}
    \end{subfigure}

    \caption{Three-way interaction profiles for VQ-Diffusion, plotting LPIPS diversity across all 8 experimental conditions. Across all facets, the lines are nearly parallel, indicating a lack of significant interaction effects. (a) Lines are almost flat, showing negligible effect of paraphrasing regardless of sampling or codebook state. (b, c) Lines show a minimal, consistent decrease from stochastic to deterministic sampling, confirming low $H_{exec}$ and absence of crossover interaction. The large gap between blue/purple lines (Full Codebook) and teal/pink lines (Subset Codebook) confirms the main effect of $H(Z)$.}
    \label{fig:interactions_vqdiff}
\end{figure*}

%% file: figs/llamagen_intersection.tex
\begin{figure}[tp]
    \centering
    \includegraphics[width=\linewidth]{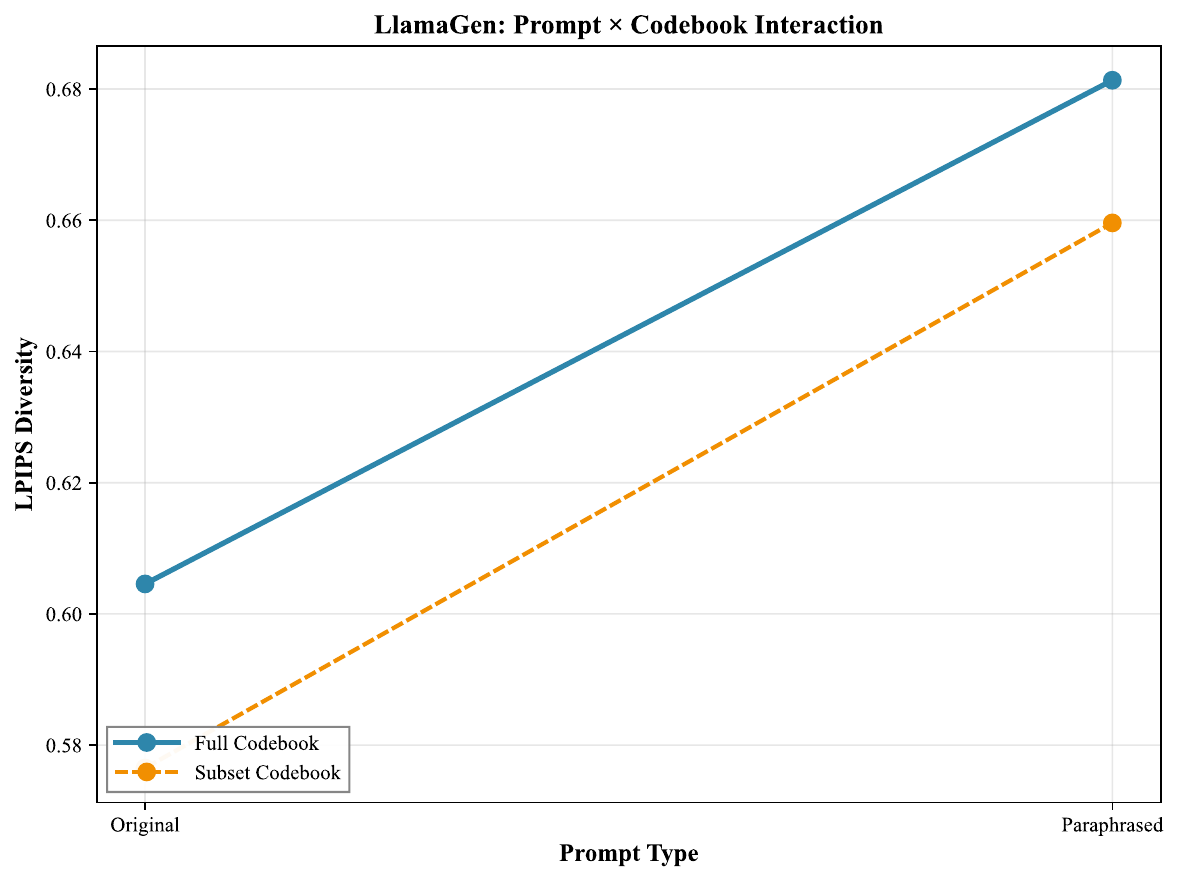}
    \caption{Two-way interaction profile for LlamaGen (stochastic sampling implied). LPIPS diversity is plotted against Prompt Type for both Full and Subset codebook states. Both lines show a positive slope, confirming the main effect of paraphrasing ($H(Z|X)$). The lines are \textbf{nearly parallel}, indicating \textbf{negligible interaction}: the benefit of paraphrasing is largely consistent across both full and subset codebook ($H(Z)$) conditions.}
    \label{fig:interaction_llamagen}
\end{figure}